\documentclass{article}

\usepackage[final]{neurips_2025}

\usepackage[utf8]{inputenc} 
\usepackage[T1]{fontenc}    
\usepackage{hyperref}       
\usepackage{url}            
\usepackage{booktabs}       
\usepackage{amsfonts}       
\usepackage{nicefrac}       
\usepackage{microtype}      
\usepackage{xcolor}    
\usepackage{booktabs}  
\usepackage{graphicx}  
\usepackage{subcaption}
\usepackage{booktabs}
\usepackage{caption}
\usepackage{tabularx}
\usepackage{multirow}
\usepackage{enumitem}
\usepackage{amsmath}
\usepackage{amssymb}
\usepackage{algorithm}
\usepackage{algorithmic}
\newcommand{\methodname}{{\tt{GPR-NIAM}}}
\PassOptionsToPackage{numbers, compress}{natbib}

\title{Global Prompt Refinement with Non-Interfering Attention Masking for One-Shot Federated Learning
}

\author{
  \textbf{Zhuang Qi}$^{1}$,
  \textbf{Yu Pan}$^{1}$,
  \textbf{Lei Meng}$^{1,}$\thanks{Corresponding author}\ ,
  \textbf{Sijin Zhou}$^{2}$,
  \textbf{Han Yu}$^{3}$,
  \textbf{Xiaoxiao Li}$^{4,5}$,
  \textbf{Xiangxu Meng}$^{1}$\\
  $^{1}$School of Software, Shandong University, China \\
  $^{2}$AIM Lab, Faculty of Engineering, Monash University, Clayton, VIC, Australia \\  
  $^{3}$College of Computing and Data Science, Nanyang Technological University, Singapore \\
  $^{4}$Department of Electrical and Computer Engineering, University of British Columbia, Canada \\
  $^5$Vector Institute, Canada\\
  \texttt{z\_qi@mail.sdu.edu.cn, 202100300334@mail.sdu.edu.cn, lmeng@sdu.edu.cn,} \\
\texttt{sjzhou1995@gmail.com,han.yu@ntu.edu.sg, xiaoxiao.li@ece.ubc.ca, mxx@sdu.edu.cn}
}

\begin{document}

\maketitle

\begin{abstract}
Federated Prompt Learning (FPL) enables communication-efficient adaptation by tuning lightweight prompts on top of frozen pre-trained models. Existing FPL methods typically rely on global information, which is only available after the second training round, to facilitate collaboration among client models. Therefore, they are inherently dependent on multi-round communication to fully exhibit their strengths. Moreover, existing one-shot federated learning methods typically focus on fitting seen tasks, but lack cross-task generalization. To bridge this gap, we propose the \underline{G}lobal \underline{P}rompt \underline{R}efinement with \underline{N}on-\underline{I}nterfering \underline{A}ttention \underline{M}asking (\methodname{}) method for one-shot FPL. The core idea is to design a masking mechanism that restricts excessive interaction between the original text embeddings and the learnable prompt embeddings. \methodname{} achieves this through the collaboration of two key modules. Firstly, the attention isolation module suppresses attention from the learnable prompt tokens to the original text tokens, and reweights the reverse attention which preserves generalization across tasks. Secondly, the cross-silo collaborative refinement module integrates decentralized visual knowledge into a unified base and calibrates the global prompt through multi-source cross-modal knowledge alignment, further mitigating the inconsistency caused by data heterogeneity. Extensive experiments conducted on ten benchmark datasets under two tasks show that \methodname{} outperforms eight state-of-the-art methods in both class-level and domain-level generalization.

\end{abstract}

\section{Introduction}
Federated learning (FL) enables collaborative modeling across data sources in a privacy-preserving manner, which has gained traction across multiple domains \cite{qi2024attentive,cai2024fgad,guo2024sample,li2020federated,fu2025less}. It learns global knowledge by aggregating model weights or gradients instead of moving data to a central entity \cite{cai2024lgfgad,9820618,yi2024federated,yi2023pfedes}. Recently, Federated Prompt Learning (FPL) has emerged to support parameter-efficient fine-tuning in this setting. It not only leverages pre-trained models (e.g., CLIP \cite{radford2021learning}) more effectively, but also reduces communication overhead between FL clients and the FL server by transmitting only learnable prompts \cite{qi2025federated,zhao2023fedprompt,ren2025advances,fan2025ten}. However, heterogeneous data distributions across clients often degrade model aggregation performance and slow down convergence \cite{yang2023efficient,guo2025federated}. This is because significant differences in the update directions across local models can cause conflicts during aggregation \cite{pan2024federated,huang2024federated,dong2022federated_FCIL,fu2025federated,zhang2024enabling}. As a result, existing FPL methods still require multi-round communications between clients and the server to reconcile inconsistencies arising from client-specific training.

To deal with data heterogeneity, early FPL solutions (e.g., PromptFoilo \cite{pan2024federated} and DP-FPL \cite{tran2025privacy}) typically focused on incorporating global prompt information to constrain the update direction of local prompts in order to enhance the consistency and stability of federated optimization \cite{pan2024federated,cui2024harmonizing,tran2025privacy,li2024global}. However, since global prompt information is typically available only after the second communication round, these methods inherently rely on multi-round communications. This poses a challenge in real-world scenarios with limited communication resources. To this end, one-shot federated learning (OSFL), which requires only a single round of client-server interaction, offers a promising direction for further exploration \cite{zhang2022dense,zeng2024one,diao2023towards,yang2024exploring,11091509}. OSFL significantly reduces the communication overhead by eliminating iterative exchanges, making it attractive for real-world deployments with strict latency or bandwidth constraints \cite{yao2025fedmho,ma2025geometric,hu2024fedmut}. However, directly applying existing OSFL methods to FPL settings is not suitable, as they are typically optimized for seen tasks while neglecting the preservation of generalization to unseen tasks.

\begin{figure}[t]
\centering
\includegraphics[width=1.0\linewidth]{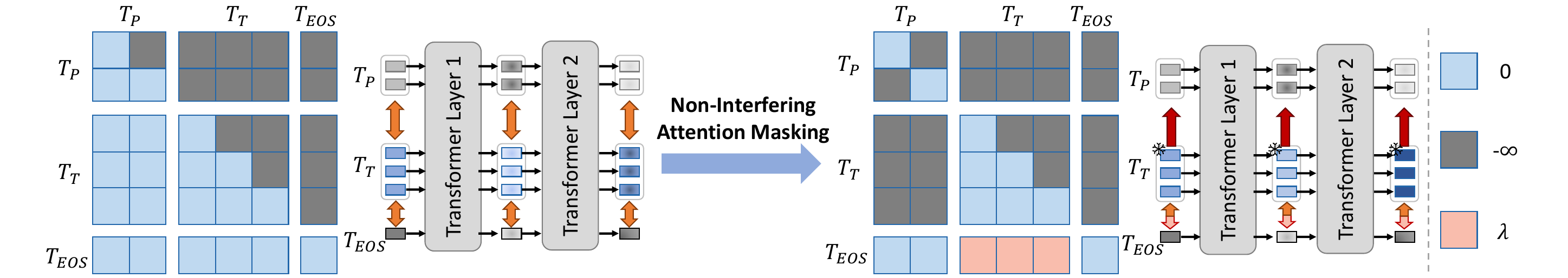}
\caption{An overview of the proposed \methodname{} method. It supports three interaction types: full attention (0), hard masking ($-\infty$), and reweighting ($\lambda$). Hard masking blocks learnable tokens $T_P$ from affecting original text tokens $T_T$, while $\lambda$-weighted links enable partial influence from special tokens $T_{EOS}$. The upper triangular mask follows causal attention for autoregressive decoding.} 
\label{fig1}
\end{figure}

To bridge this important gap, we propose the \underline{G}lobal \underline{P}rompt \underline{R}efinement with \underline{N}on-\underline{I}nterfering \underline{A}ttention \underline{M}asking (\methodname{}) method. It prevents the modification of the original token embeddings through a masking mechanism, thereby preserving transferable knowledge across tasks. As shown in Figure \ref{fig1}, the interaction between the learnable prompt embeddings and the original text token embeddings is unidirectional, which ensures that the latter remain unchanged. Specifically, \methodname{} consists of two main modules. Firstly, the Attention Isolation module selectively regulates the flow of prompt information through a novel masking strategy. It suppresses attention from the learnable prompt tokens to the original text tokens across all encoder layers, thereby effectively preventing alterations to the textual input token embeddings. In addition, it re-weights the reverse attention to encourage the model to extract task-relevant features more effectively from the original texts. Secondly, the cross-silo collaborative refinement (CSCR) module leverages visual representations extracted from multiple clients and utilizes multi-source visual supervision to  refine the global prompt in a centralized manner. This further mitigates the negative impact of data heterogeneity.

We conduct extensive experiments on 10 datasets under two tasks: 1) base-to-base/novel generalization and 2) leave-one-domain-out generalization, and carry out performance comparisons, ablation studies, in-depth analysis and case studies. The results demonstrate that \methodname{} achieves superior overall performance by maintaining a strong balance between fitting seen tasks and generalizing to unseen ones, significantly outperforming eight competitive state-of-the-art approaches.

\section{Related Work}

\subsection{Federated Prompt Learning}
FPL aims to integrate FL with prompt learning by freezing the backbone model within the FL framework and introducing a set of learnable prompt parameters \cite{pan2024federated,cui2024harmonizing,tran2025privacy,li2024global,su2024federated,jia2025llm,guo2023pfedprompt,deng2024unlocking}. This enables efficient personalization for on-device data while significantly reducing communication overhead, as only prompt parameters are shared between clients and the server. To mitigate the challenges posed by data heterogeneity across clients, popular FPL methods (e.g., FedOTP \cite{li2024global}, FedPGP \cite{cui2024harmonizing}, PromptFolio \cite{pan2024federated}, DP-FPL \cite{tran2025privacy}) mostly adopt a shared prompt paradigm. In this design, a global prompt is constructed on the server and combined with client-specific information to improve training stability and generalization under non-IID distributions. However, these methods are essentially designed for multi-round communications and require access to global feedback, which limits their effectiveness in one-shot FL scenarios.

\subsection{One-Shot Federated Learning}
OSFL is a communication-efficient variant of federated learning, where model training is completed within a single round of client-server interaction \cite{zhang2022dense,zeng2024one,li2025feature,diao2023towards,yang2024exploring,yang2024feddeo,allouah2024revisiting,chiang2023optimal,heinbaugh2023data,wang2024one}. Recent studies on OSFL mainly focus on generative and ensemble-based approaches. The former employs generative models to synthesize proxy data or representations on the server. For example, FedCVAE \cite{heinbaugh2023data} uses a variational autoencoder to compress local knowledge. FedDISC \cite{yang2024exploring} and FedDEO \cite{yang2024feddeo} leverage diffusion models to generate synthetic training data. The latter is dedicated to refining the fusion of model predictions from different clients. For example, FedOV \cite{diao2023towards} introduces an open-set recognition-based voting mechanism to address label skew in OSFL; IntactOFL \cite{zeng2024one} learns weighted or expert-based combinations to enhance overall performance. However, these methods are only designed for conventional tasks, aiming to fit the training objective while overlooking the preservation of cross-task generalization.

\section{Preliminaries}

Suppose there are $K$ clients denoted by $\mathcal{C} = \{C_1, C_2, \dots, C_K\}$, and each client $C_k \in \mathcal{C}$ holds a private dataset $\mathcal{D}_k$. The goal of traditional FL is to collaboratively train a global model $f$ by minimizing the objective: $\min_{f} \sum_{k=1}^K p_k \cdot \mathcal{L}(f, \mathcal{D}_k)$, where $p_k = \frac{|\mathcal{D}_k|}{\sum_{j=1}^K |\mathcal{D}_j|}$ is the weight of client $k$, and $\mathcal{L}(\cdot)$ is a task-specific loss function \cite{hu2024fedcross,qi2025global,li2025re,zhangcausality}. In FPL, each client $k$ optimizes a set of learnable prompts $\delta_k$ using its private dataset $\mathcal{D}_k$, while keeping the backbone model $f$ frozen: $\min_{\delta_k} \mathcal{L}(f; \delta_k, \mathcal{D}_k)$. To support class-level prediction, the prompt for class $c$ is constructed by concatenating the learnable prompt and a fixed class token. Specifically, let $\mathbf{p}_c \in \mathbb{R}^{N_t \times d_t}$ denote the class token for class $c$, and $\delta_k \in \mathbb{R}^{N_p \times d_t}$ denote the learnable prompt from client $k$, where $N_t$ and $N_p$ are the lengths of the fixed and learnable prompts, $d_t$ is the dimension of hidden embedding, respectively. The full prompt can be formed as $P_c^{(k)} = [\delta_k; \mathbf{p}_c]$. The text feature for class $c$ is computed by feeding the prompt into a text encoder $\mathcal{T}(\cdot)$: $\mathbf{f}_t^c = \mathcal{T}(P_c^{(k)})$. Given an input image $\mathbf{x}$, the image feature is extracted by the frozen image encoder $f$: $\mathbf{f}_v = f(\mathbf{x})$. The similarity score between the image and class-$c$ textual features is calculated as $\rho_c = \operatorname{Cosine}(\mathbf{f}_t^c, \mathbf{f}_v)$, and the prediction probability is computed via softmax over similarities: $p(\hat{y}=c \mid \mathbf{x}) = \frac{\exp(\rho_c / \tau)}{\sum_{j=1}^K \exp(\rho_j / \tau)}$. The training objective is the cross-entropy loss between the similarity-based prediction and the ground-truth label: $\mathcal{L}_{\text{ce}} = \ell(\boldsymbol{\rho}, \mathbf{e}_y)$, where $\boldsymbol{\rho} = [\rho_1, \dots, \rho_K]$, $\mathbf{e}_y$ is the one-hot label vector, and $\ell$ denotes the cross-entropy loss function. After local training, client $C_k$ uploads its learned prompt $\delta_k$ to the server. The server then aggregates the local prompts $\{\delta_k\}_{k=1}^K$ to construct a global prompt $\delta_g$, typically by weighted averaging: $\delta_g = \sum_{k=1}^K p_k \cdot \delta_k$.

\section{The Proposed \methodname{} Method}
\begin{figure}[t]
\centering
\includegraphics[width=1.0\linewidth]{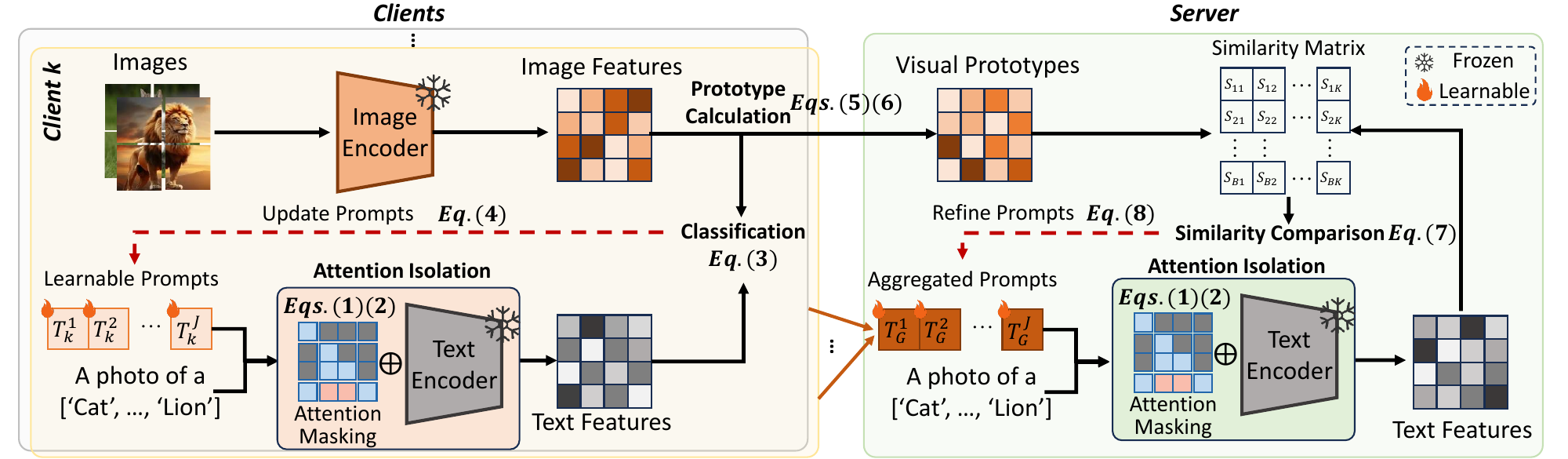}
\caption{Illustration of the \methodname{} framework. The Attention Isolation module preserves the original text embeddings to reduce the forgetting of generalized knowledge. The global prompt is refined using visual prototypes from all clients to mitigate data heterogeneity.
} 
\label{fig2}
\end{figure}

In this section, we propose the non-interfering attention masking method for one-shot federated prompt learning, called \methodname{}. Its
core idea is to preserve the transferable knowledge by reducing the influence of learnable prompt tokens on the input text tokens, as illustrated in Figure \ref{fig2}.
 
\subsection{The Attention Isolation (AttnIso) Module}
Prompt tuning typically focuses on fitting local training tasks. However, data heterogeneity often leads to divergent optimization across clients and weakens its cross-task generalization. To address this, the AttnIso module designs attention masks to selectively block the information flow from learnable prompt tokens to original text tokens (e.g. ``a photo of a [CLASS]''). We denote the text attention matrix as $M\in\mathbb{R}^{(1+N_p+N_t)*(1+N_p+N_t)}$, where $N_p$ is the number of learnable prompts, $N_t$ is the length of input textual feature tokens. The additional 1 corresponds to the [EOS] token. In addition, we apply customized attention control for the [EOS] token to ensure proper aggregation of contextual information from both the prompt and the input tokens.

To achieve the above goal, the text attention matrix $M$ can be formulated as follows, 
\begin{equation}
M^{i,j} =
\begin{cases}
-\infty, &
\begin{aligned}
& \text{if } i \in [1, N_p] \land j \in [N_p + 1, N_p + N_t] \quad &  \\
& \text{or } i \in [N_p + 1, N_p + N_t] \land j \in [1, N_p] \quad &  \\
& \text{or } j = N_p + N_t + 1 \quad & 
\end{aligned} \quad \text{(Hard Masking)}\\
\lambda, & \text{if } i = N_p + N_t + 1 \land j \in [N_p + 1, N_p + N_t] \quad \text{(Reweighting)}  \\
0, & \text{otherwise} \quad \quad \quad \quad \quad \quad \quad \quad \quad \quad \quad \quad \quad \quad \quad   \text{(Full Attention)}
\end{cases}
\end{equation}
where $M^{i,j}$ denotes the $i$-th row and $j$-th column element of $M$, and $\land$ denotes logical AND. Moreover, to integrate the attention mask $M$ into the text encoder $\mathcal{T}$, we replace the original computation with the following:
\begin{equation}
\resizebox{0.91\hsize}{!}{$
\begin{aligned}
\mathbf{y}^{(l+1)} = \mathcal{T}_{l+1}(\mathbf{y}^{(l)}, &M) = \text{Concat}(h_1, h_2, \dots, h_H)W^O = [\tilde{p}_{l+1}^1, \dots, \tilde{p}_{l+1}^{N_p}, \tilde{t}_{l+1}^1, \dots, \tilde{t}_{l+1}^{N_t}, \tilde{s}_{l+1}^{\text{eos}}], \\
&\text{where } h_i = \operatorname{softmax}\left(\frac{Q_i K_i^T}{\sqrt{d_k}} + M\right)V_i, \quad i = 1, \dots, H
\end{aligned}$}
\end{equation}
where $\mathbf{y}^{(l)} \in \mathbb{R}^{(N_p + N_t + 1) \times d_t}$ and $\mathbf{y}^{(l+1)} \in \mathbb{R}^{(N_p + N_t + 1) \times d_t}$ are the input and output of the $(l+1)$-th encoder layer. $\tilde{p}_{l+1}^i$, $\tilde{t}_{l+1}^j$, and $\tilde{s}_{l+1}^{\text{eos}}$ represent the output embeddings of the $i$-th prompt token, the $j$-th text token, and the [EOS] token, respectively.
$Q_i, K_i, V_i \in \mathbb{R}^{(N_p + N_t + 1) \times d_k}$ are the query, key, and value matrices for the $i$-th attention head, $d_k=d_t/H$ denotes the hidden dimension of each attention head. 
$\text{Concat}(h_1, \dots, h_H) \in \mathbb{R}^{(N_p + N_t + 1) \times H d_k}$ denotes the concatenated outputs of all heads, and $ W^O \in \mathbb{R}^{H d_k \times d_t}$ is the output projection matrix mapping the result back to $d_t$. Following the masked encoding process, we obtain the final output of the text encoder as: $
\mathbf{y}^{(L)} = [\tilde{p}_{L}^1, \dots, \tilde{p}_{L}^{N_p}, \tilde{t}_{L}^1, \dots, \tilde{t}_{L}^{N_t}, \tilde{s}_{L}^{\text{eos}}]
$. Afterwards, $\tilde{s}_{L}^{\text{eos}}$ is projected into a shared embedding space via a pre-trained head $H_t(\cdot)$. On the visual side, the input image $\mathbf{x}$ is encoded by a frozen vision encoder $f(\cdot)$ to obtain the visual representation. Overall, it focuses on minimizing the objective:
\begin{equation}
\resizebox{0.93\hsize}{!}{$
\min_{\delta_1, \dots, \delta_K} \sum_{k=1}^{K} \frac{|\mathcal{D}_k|}{|\mathcal{D}|}
\cdot \mathbb{E}_{(\mathbf{x}, y) \sim \mathcal{D}_k}
\left[
- \log
\left(
\frac{
\exp\left( \operatorname{sim}\left(f_v(\mathbf{x}), H_t\left( \tilde{s}^{(y)}_{L,k} \right)\right) \right)
}{
\sum_{c=1}^{C} \exp\left( \operatorname{sim}\left(f_v(\mathbf{x}), H_t\left( \tilde{s}^{(c)}_{L,k} \right)\right) \right)
}
\right)
\right],$}
\end{equation}
where $\tilde{s}^{(c)}_{L,k}$ denotes client $k$’s [EOS] embedding for class $c$, $\operatorname{sim}(\cdot,\cdot)$ denotes cosine similarity function. The learnable prompts can be optimized by gradient descent in all clients:
\begin{equation}\label{eq4}
\resizebox{0.93\hsize}{!}{$
\delta_k \leftarrow \delta_k - \eta \cdot \frac{1}{|\mathcal{B}_k|} \sum_{(\mathbf{x}, y) \in \mathcal{B}_k}
\nabla_{\delta_k}
\left[
- \log
\left(
\frac{
\exp\left( \operatorname{sim}\left(f_v(\mathbf{x}), H_t\left( \tilde{s}^{(y)}_{L,k} \right)\right) \right)
}{
\sum_{c=1}^{C} \exp\left( \operatorname{sim}\left(f_v(\mathbf{x}), H_t\left( \tilde{s}^{(c)}_{L,k} \right)\right) \right)
}
\right)
\right],
\quad \forall k \in [K],$}
\end{equation}
where $\eta$ denotes the learning rate, $\mathcal{B}_k$ is a batch data.

\subsection{Cross-Silo Collaborative Refinement (CSCR) Module}
To further alleviate the inconsistency caused by data heterogeneity, the CSCR module calibrates the global prompt through prototype-guided refinement across clients. To support this, each client $C_k$ extracts class-wise visual prototypes from its local dataset $\mathcal{D}_k$. For class $c$, the prototype is computed as a randomly weighted average of the feature vectors from local samples in that class:
\begin{equation}\label{eq5}
   f_{v,k}^c = \sum_{\mathbf{x} \in \mathcal{D}_k^c} \alpha_{\mathbf{x}} \cdot f_v(\mathbf{x}),
\quad \text{where } \sum_{\mathbf{x} \in \mathcal{D}_k^c} \alpha_{\mathbf{x}} = 1,\; \alpha_{\mathbf{x}} \sim \mathcal{U}(0,1). 
\end{equation}
where $\alpha_{\mathbf{x}}$ denotes a random weight for each sample $\mathbf{x}$ in class $c$, drawn from a uniform distribution and normalized to ensure the weights sum to 1. Notably, we can randomly sample the weight vector $n$ times to generate multiple diverse prototypes per class $\mathbf{P}_k^c = \left[ f_{v,k}^{c,1}, f_{v,k}^{c,2}, \dots, f_{v,k}^{c,n} \right]$, which enhances representational diversity, which differs from existing methods \cite{qi2023cross,meng2024improving,qi2024cross,meng2025robust,qi2025cross}. $f_{v,k}^{c,i}$ denotes $i$-th visual prototypes of class $c$ in client $k$. Afterwards, each client $C_k$ sends $\mathbf{P}_k = \{\mathbf{P}_k^1, \dots, \mathbf{P}_k^N\}$ to the server, which aggregates all received prototypes into a global pool $\mathcal{P}$:
\begin{equation}\label{eq6}
\mathcal{P} = \bigcup_{k=1}^{K} \mathbf{P}_k = \left\{ \mathbf{P}_{k}^{c} \mid k \in [K],\; c \in [n] \right\},
\end{equation}
To calibrate the global prompt $\delta_g$, we leverage the global prototype pool $\mathcal{P} = \{\mathcal{P}^1, \dots, \mathcal{P}^C\}$ and enforce the alignment between textual representations and visual prototypes via a class-level contrastive objective. Specifically, we design the following loss function:
\begin{equation}
\resizebox{0.94\hsize}{!}{$
\min_{\delta_g}
- \frac{1}{|\mathcal{P}|}
\sum_{f \in \mathcal{P}}
\log \sigma\left(
\operatorname{sim}\left(f,\; H_t(\mathcal{T}(\delta_g, \text{t}_c))\right) -
\log \sum_{j=1}^{C} \exp\left(
\operatorname{sim}\left(f,\; H_t(\mathcal{T}(\delta_g, \text{t}_j))\right)
\right)
\right),$}
\end{equation}
\noindent
where $\sigma(\cdot)$ is the sigmoid function, $\text{t}_c$ denotes 'a photo of c-th classname'. We then follow the gradient descent method to obtain the optimal parameters: 
\begin{equation}\label{eq8}
\resizebox{0.94\hsize}{!}{$
\delta_g \leftarrow \delta_g - \eta \cdot \frac{1}{|\mathcal{P}|} \sum_{(f, c) \in \mathcal{P}} 
\nabla_{\delta_g} \log \sigma\left(
\operatorname{sim}\left(f,\; H_t(\mathcal{T}(\delta_g, \text{t}_c))\right) -
\log \sum_{j=1}^{C} \exp\left(
\operatorname{sim}\left(f,\; H_t(\mathcal{T}(\delta_g, \text{t}_j))\right)
\right)
\right),$}
\end{equation}
where the global prompt $\delta_g$ is initialized as the weighted average of all client-specific prompts (i.e., $\delta_g \leftarrow \sum_{k=1}^K \frac{|\mathcal{D}_k|}{|\mathcal{D}|} \cdot \delta_k$), and $\eta$ is a learning rate.

\subsection{Training Procedure}
In this section, we provide the detailed training procedure of \methodname{} in Algorithm~\ref{alg1}. As a two-stage optimization strategy, \methodname{} consists of a local prompt tuning phase and a global prompt refinement phase. Notably, it can be easily integrated into multiple prompt learning methods such as CoOp and TCP. In the first stage, each client initializes its local prompt from the global prompt and updates it on private data with masked attention learning. Simultaneously, class-wise visual prototypes are obtained by performing randomly weighted aggregation over local feature representations. Subsequently, the global prompt is optimized to align with all class prototypes in this pool via cross-silo collaborative retraining. With the two-stage design, \methodname{} alleviates the challenges posed by heterogeneous client data and promotes the retention of transferable knowledge for unseen tasks. 

Moreover, the local prompt tuning stage is optional. Firstly, this stage provides a more meaningful initialization for the global prompt by adapting it to all clients data. Secondly, it generates class-wise visual prototypes to guide global refinement. While the framework can still perform global refinement without this phase, since it does not affect prototype modeling, omitting it might result in less effective initialization and slightly limit the overall refinement performance.

\begin{algorithm}[t]
\caption{\textsc{GPR-NIAM}}
\label{alg1}
\begin{algorithmic}[1]

\STATE \textbf{Initialize} the global prompt parameter $\delta^0$
\STATE \textbf{// Local Prompt Tuning Stage}
\FOR{each client $C_k \in \mathcal{C}$ in parallel}
    \STATE Initialize local prompt parameter $\delta_k = \delta^{0}$
    \FOR{$e = 1, \dots, E$}
        \STATE Sample a mini-batch $\zeta$ from local data $\mathcal{D}_k$  
        \STATE Update local prompt $\delta_k$ using masked attention by Eq.~\eqref{eq4}
    \ENDFOR
    \STATE Compute visual prototype features for each class using Eq.~\eqref{eq5}
\ENDFOR
\STATE \textbf{// Global Prompt Refinement Stage}
\STATE Aggregate all visual prototypes to form a global prototype pool $\mathcal{P}$ by Eq.~\eqref{eq6}
\FOR{$e' = 1, \dots, E^{'}$}
    \STATE Sample a mini-batch of prototypes $\zeta^{'}$ from $\mathcal{P}$  
    \STATE Update global prompt $\delta_g$ via prototype-guided alignment using Eq.~\eqref{eq8}
\ENDFOR
\end{algorithmic}
\end{algorithm}

\section{Experimental Evaluation}
\subsection{Experiment Settings}
\paragraph{Datasets}
Following existing works \cite{pan2024federated,qiu2024federated}, experiments are conducted on CIFAR10 \cite{krizhevsky2009learning}, OxfordPets \cite{parkhi2012cats}, Caltech101 \cite{fei2004learning}, DTD \cite{cimpoi2014describing}, FGVCAircraft \cite{maji2013fine}, Flowers102 \cite{nilsback2008automated}, StanfordCars \cite{krause20133d}, and UCF101 \cite{soomro2012ucf101} to evaluate base-to-base/novel generalization, and on Office-Home \cite{torralba2011unbiased} and DomainNet \cite{peng2019moment} to assess leave-one-domain-out generalization. Their statistics are shown in Table \ref{tab4}.

\paragraph{Evaluation Metrics}
Following studies~\cite{qi2023cross,qi2024cross,liang2025tta}, we report the Top-1 Accuracy ($\text{Acc}$) results. For base-to-base/novel generalization tasks, we also compute the harmonic mean ($\text{HM} = \frac{2 \cdot \text{Acc}_{\text{base}} \cdot \text{Acc}_{\text{novel}}}{\text{Acc}_{\text{base}} + \text{Acc}_{\text{novel}}}$) of accuracy on base ($\text{Acc}_{\text{base}}$) and novel classes ($\text{Acc}_{\text{novel}}$) to evaluate overall performance.

\begin{table*}[!b]
\centering
\caption{Performance comparison between \methodname{} and baselines on 8 datasets with $\beta=0.5$ in the base-to-base/novel generalization cases.}
\renewcommand{\arraystretch}{1} 
\resizebox{0.9\textwidth}{!}{
\begin{tabular}{lccc|lccc|lccc}
\toprule
\multicolumn{4}{c|}{\textbf{(a) Average over 8 datasets}} & 
\multicolumn{4}{c|}{\textbf{(b) CIFAR10}} & 
\multicolumn{4}{c}{\textbf{(c) OxfordPets}} \\\hline
\textbf{Method} & \textbf{Base} & \textbf{Novel} & \textbf{HM} & 
\textbf{Method} & \textbf{Base} & \textbf{Novel} & \textbf{HM} & 
\textbf{Method} & \textbf{Base} & \textbf{Novel} & \textbf{HM} \\
\midrule
CLIP        & 65.79 & 74.32 & 69.70 & CLIP      & 95.30 & 95.50 & 95.39 & CLIP        & 81.53 & 95.33 & 87.89 \\
PromptFL    & 72.57 & 72.80 & 72.64 & PromptFL    & 95.14 & 96.26 & 95.69 & PromptFL    & 91.31 & 93.70 & 92.48 \\
RPO & 70.52 & 73.63 & 71.99 & RPO & 95.68 & 94.74 & 95.20 & RPO & 89.00 & 95.49 & 92.13 \\
FedTPG      & 70.46 & 73.09 & 71.69 & FedTPG      & 95.38 & 96.56 & 95.96 & FedTPG      & 91.66 & 95.13 & 93.36 \\
TCP         & 72.46 & 74.60 & 73.41 & TCP         & 95.68 & 96.52 & 96.09 & TCP         & 90.01 & 95.39 & 92.62 \\
PromptFolio & 73.25 & 73.35 & 73.15 & PromptFolio & \textbf{96.04} & 96.16 & 96.06 & PromptFolio & 91.13 & \textbf{96.10} & 93.54 \\
DP-FPL & 71.90 & 71.75 & 69.88 & DP-FPL & 95.42 & 95.18 & 95.29 & DP-FPL & 91.97 & 94.80 & 93.36 \\
FedDISC     & 56.43 & 54.84 & 53.87 & FedDISC     & 94.30 & 92.82 & 93.55 & FedDISC     & 83.83 & 72.07 & 77.50 \\
FedELMY      & \textbf{76.62} & 48.23 & 60.03 & FedELMY      & 95.32 & 40.00 & 79.66 & FedELMY      & 89.28 & 76.47 & 82.38 \\\hline
\methodname{}$_{P}$      & 75.56 & 73.28 & 74.24 & \methodname{}$_{P}$      & 95.66 & \textbf{97.00} & \textbf{96.32} & \methodname{}$_{P}$      & 91.18 & 95.89 & 93.47 \\
\methodname{}$_{T}$      & 76.43 & \textbf{74.48} & \textbf{75.30} & \methodname{}$_{T}$      & 95.84 & 96.48 & 96.15 & \methodname{}$_{T}$      & \textbf{92.48} & 95.14 & \textbf{93.79} \\
\bottomrule
\multicolumn{4}{c|}{\textbf{(d) Caltech101}} & 
\multicolumn{4}{c|}{\textbf{(e) DTD}} & 
\multicolumn{4}{c}{\textbf{(f) FGVCAircraft}} \\\hline
\textbf{Method} & \textbf{Base} & \textbf{Novel} & \textbf{HM} & 
\textbf{Method} & \textbf{Base} & \textbf{Novel} & \textbf{HM} & 
\textbf{Method} & \textbf{Base} & \textbf{Novel} & \textbf{HM} \\
\midrule
CLIP        & 82.27 & 94.05 & 87.77 & CLIP        & 53.35 & \textbf{59.05} & 56.06 & CLIP        & 26.51 & 31.21 & 28.67 \\
PromptFL    & 88.27 & 93.88 & 90.99 & PromptFL    & 60.06 & 53.86 & 56.79 & PromptFL    & 31.19 & 30.91 & 31.05 \\
RPO & 89.90 & 93.80 & 91.81 & RPO & 54.62 & 59.17 & 56.81 & RPO & 29.27 & 30.43 & 29.84 \\
FedTPG      & 89.12 & 92.61 & 90.83 & FedTPG      & 56.71 & 55.43 & 56.06 & FedTPG      & 29.57 & 30.25 & 29.90 \\
TCP         & 88.20 & \textbf{95.33} & 91.62 & TCP         & 61.68 & 55.48 & 58.41 & TCP         & 30.77 & \textbf{32.71} & 31.71 \\
PromptFolio & 87.28 & 93.12 & 90.10 & PromptFolio & 61.11 & 54.58 & 57.66 & PromptFolio & 31.37 & 32.39 & \textbf{31.87} \\
DP-FPL & 87.92 & 93.46 & 90.60 & DP-FPL & 62.26 & 49.63 & 55.24 & DP-FPL & 28.85 & 28.15 & 28.49 \\
FedDISC     & 44.75 & 85.22 & 58.68 & FedDISC     & \textbf{71.52} & 38.28 & 49.87 & FedDISC     & 20.87 & 13.20 & 16.17 \\
FedELMY      & 87.71 & 89.21 & 88.45 & FedELMY      & 70.13 & 16.18 & 26.29 & FedELMY      & \textbf{33.77} & 19.48 & 24.70 \\\hline
\methodname{}$_{P}$      & 91.59 & 94.22 & 92.89 & \methodname{}$_{P}$      & 68.40 & 53.50 & 60.04 & \methodname{}$_{P}$      & 30.95 & 29.41 & 30.16 \\
\methodname{}$_{T}$      & \textbf{92.65} & 94.73 & \textbf{93.68} & \methodname{}$_{T}$      & 69.90 & 55.91 & \textbf{62.13} & \methodname{}$_{T}$      & 32.15 & 31.12 & 31.62 \\
\bottomrule
\multicolumn{4}{c|}{\textbf{(g) Flowers102}} & 
\multicolumn{4}{c|}{\textbf{(h) StanfordCars}} & 
\multicolumn{4}{c}{\textbf{(i) UCF101}} \\\hline
\textbf{Method} & \textbf{Base} & \textbf{Novel} & \textbf{HM} & 
\textbf{Method} & \textbf{Base} & \textbf{Novel} & \textbf{HM} & 
\textbf{Method} & \textbf{Base} & \textbf{Novel} & \textbf{HM} \\
\midrule
CLIP        & 62.39 & 78.10 & 69.36 & CLIP        & 56.87 & 69.40 & 62.51 & CLIP        & 68.12 & 71.92 & 69.97 \\
PromptFL    & 79.88 & 75.15 & 77.44 & PromptFL    & 62.96 & 67.30 & 65.02 & PromptFL    & 71.82 & 71.55 & 71.68 \\
RPO & 74.92 & 76.21 & 75.56 & RPO & 58.64 & 68.73 & 63.29 & RPO & 72.19 & 70.49 & 71.33 \\
FedTPG      & 72.30 & 77.26 & 74.70 & FedTPG      & 58.67 & 69.78 & 63.74 & FedTPG      & 70.34 & 67.74 & 69.01 \\
TCP         & 77.25 & \textbf{78.73} & 77.98 & TCP         & 60.19 & 71.18 & 65.22 & TCP         & 75.93 & 71.52 & 73.65 \\
PromptFolio & 81.04 & 74.51 & 77.63 & PromptFolio & 61.22 & \textbf{69.91} & 65.27 & PromptFolio & 76.88 & 69.75 & 73.14 \\
DP-FPL & 74.63 & 77.26 & 75.92 & DP-FPL & 59.07 & 66.05 & 62.36 & DP-FPL & 75.14 & 69.49 & 72.20 \\
FedDISC     & 24.95 & 32.21 & 28.11 & FedDISC     & 42.77 & 49.92 & 46.07 & FedDISC     & 68.49 & 55.03 & 61.03 \\
FedELMY      & \textbf{89.79} & 55.15 & 68.33 & FedELMY      & \textbf{67.09} & 47.97 & 55.94 & FedELMY      & \textbf{79.94} & 41.41 & 54.56 \\\hline
\methodname{}$_{P}$      & 84.54 & 78.52 & 81.42 & \methodname{}$_{P}$      & 63.59 & 67.48 & 65.48 & \methodname{}$_{P}$      & 78.57
 & 70.28 & 74.19 \\
\methodname{}$_{T}$      & 86.88 & 77.05 & \textbf{81.67} & \methodname{}$_{T}$      & 63.54 & 69.18 & \textbf{66.24} & \methodname{}$_{T}$      & 78.04 & \textbf{76.27} & \textbf{77.14} \\
\bottomrule
\end{tabular}
}
\label{tab1}
\vspace{-0.3cm}
\end{table*}

\paragraph{Implementation Details}
In all experiments, we set local training and global refinement epochs to 10, using the SGD optimizer with a learning rate of 0.001, a weight decay of 0.001, and a batch size of 32. The number of communication rounds is 1. We simulate 50 clients for CIFAR-10 and 10 clients for other datasets in base-to-base/novel generalization, while using 9 clients for Office-Home and 15 for DomainNet. To simulate non-IID data, we adopt a Dirichlet distribution with parameter $\beta = 0.5$. Each client generates $n \in \{5, 10, 20\}$ visual prototypes per class, and the reweighting parameter $\lambda$ is selected from $\{0.2, 0.5, 0.7, 1.0\}$. More details and results can be found in the Appendix. And each client has one NVIDIA RTX 3090 with 24 GB GPU for training.

\subsection{Results and Discussion}
This section compares \methodname{}-enhanced variants (\methodname{}$_P$ and \methodname{}$_T$, derived by integrating \methodname{} into PromptFL and TCP, respectively) with nine baselines: Zero-Shot CLIP \cite{radford2021learning} with hand-crafted text prompt template, e.g., “a photo of a [class]” \cite{li2024knn}, PromptFL \cite{guo2023PromptFL}, 
RPO \cite{lee2023read}, FedTPG \cite{qiu2024federated}, TCP \cite{yao2024tcp}, PromptFolio \cite{pan2024federated}, DP-FPL \cite{tran2025privacy}, FedDISC \cite{yang2024exploring}, FedELMY \cite{wang2024one}. For fair comparison, all methods use the same backbone: a frozen CLIP with ViT-B16~\cite{radford2021learning}. Following prior work, the learnable prompt is a $10 \times 512$ parameter matrix, with 10 prompt tokens and a 512-dimensional embedding.

\textbf{Base-to-Base/Novel Generalization.} We evaluate the generalization capability of \methodname{} by comparing its performance on base and novel classes, and the overall harmonic mean (HM) across all benchmarks. The results are presented in Table \ref{tab1} and key observations are summarized as follows:
\begin{itemize}[leftmargin=10pt]
\item \methodname{} attains second-best performance on base classes among all methods, showing particular competitiveness against FedMLEY while maintaining more balanced capabilities across all classes.
\item \methodname{} yields competitive performance on novel classes. Notably, it ranks first on Flowers102 and UCF101, underscoring its strong generalization capability to unseen tasks. Moreover, FedDISC may fail on both base and novel classes due to inevitable gaps between synthetic and real samples.
\item Benefiting from the synergy of attention masking and cross-silo collaborative retraining, \methodname{} achieves the best HM scores across all benchmarks.
\end{itemize}
\begin{table*}[t]
\centering
\caption{Performance comparison on Office-Home and DomainNet under the leave-one-domain-out setting, with each domain's data randomly split across 3 clients.}
\renewcommand{\arraystretch}{1.0}
\resizebox{\textwidth}{!}{
\begin{tabular}{l|ccccc|cccccccc}
\toprule
\multirow{2}{*}{\textbf{Method}} & 
\multicolumn{5}{c|}{\textbf{Office-Home}}  & 
\multicolumn{7}{c}{\textbf{DomainNet}} \\
\cmidrule(lr){2-6} \cmidrule(lr){7-13}
& Art & Clipart & Product & RealWorld & Average & 
Clipart & Infograph & Painting & Quickdraw & Real & Sketch & Average\\
\midrule
CLIP      & 74.57 & 62.33 & 80.17 & 79.38 & 74.08 & 92.87 & 77.02 & 85.35 & 43.06 & 98.61 & 94.22 & 81.85 \\
PromptFL      & 74.24 & 66.20 & 82.47 & 79.77 & 75.17 & 93.87 & 80.82 & 87.76 & 49.47 & 98.72 & 94.82 & 84.24 \\
RPO     & 74.86 & 64.33 & 84.31 & 80.74 & 76.06 & 93.53 & 78.32 & 86.15 & 49.35 & 98.83 & 94.87 & 83.50 \\
FedTPG       & 74.04 & 64.55 & 83.32 & 80.01 & 75.48 & \textbf{95.14} & 80.93 & 88.25 & 52.89 & 98.97 & 95.40 & 85.26 \\
TCP       & 75.31 & 66.20 & 83.17 & 80.97 & 76.41 & 94.13 & 79.80 & 88.05 & 45.85 & 98.80 & 95.34 & 83.66 \\
PromptFolio     & 77.21 & 67.74 & 84.74 & 81.98 & 77.92 & 94.87 & \textbf{81.19} & 88.61 & 47.59 & 98.85 & 95.02 & 84.35 \\
DP-FPL   & 75.03 & 65.91 & 83.71 & 79.82 & 76.12 & 95.03 & 78.41 & 86.74 & 54.03 & 98.83 & 94.85 & 84.64 \\
FedDISC     & 72.27 & 65.40 & 81.77 & 79.11 & 74.63 & 93.82 & 80.42 & 86.17 & 43.74 & 98.66 & 94.05 & 82.81 \\
FedELMY   & 75.11 & 66.34 & 84.12 & 79.69 & 76.31 & 94.21 & 77.65 & 86.81 & 48.78 & 98.83 & 94.88 & 83.53 \\\hline
\methodname{}$_P$     & 78.32 & 71.40 & 87.61 & 84.96 & 80.57 & 94.72 & 80.13 & \textbf{88.70} & \textbf{56.98} & \textbf{99.18} & 94.88 & 85.76 \\
\methodname{}$_T$      & \textbf{79.39} & \textbf{72.06} & \textbf{88.03} & \textbf{85.49} & \textbf{81.24} & 95.03 & 80.82 & 87.92 & 56.32 & 99.08 & \textbf{95.60} & \textbf{85.79} \\
\bottomrule
\end{tabular}
}
\label{tab2}
\end{table*}

\textbf{Leave-One-Domain-Out Generalization.} Table~\ref{tab2} presents the domain generalization results under a leave-one-domain-out setting. The main findings are summarized below:
\begin{itemize}[leftmargin=10pt]
\item \methodname{} consistently achieve the highest average accuracy across most domains. Notably, \methodname{}$_{T}$ ranks first in 9 out of 12 domains, highlighting its strength in cross-domain generalization, especially in challenging cases like Infograph and Quickdraw.
\item By combining token-level and client-level perspectives, \methodname{} outperforms single-view methods such as PromptFL and FedTPG, which benefits from attention masking for preserving original semantics and a unified global prompt for cross-domain knowledge integration.
\end{itemize}

\subsection{Ablation Study}
This section presents an ablation study to examine the effects of key components in \methodname{}, including the Attention Isolation (AttnIso) module and the Cross-Silo Collaborative Retraining (CSCR) module. The results are reported in Table~\ref{tab3}.

\begin{itemize}[leftmargin=10pt]
\item Different configurations of the AttnIso module exhibit distinct strengths. Hard masking (AttnIso$_{\text{HA}}$) effectively suppresses token interference and enhances novel-class generalization, while reweighting (AttnIso$_{\text{RE}}$) provides a more stable improvement. Their combination leads to improved overall generalization, showing that the two strategies are complementary.
\item The CSCR module plays a vital role in optimizing base-class performance by learning a unified prompt that mitigates inter-client discrepancies, which can alleviate the negative impact of data heterogeneity. However, its contribution to novel-class generalization remains limited due to insufficient adaptability to unseen categories.
\item By integrating both modules, \methodname{} achieves an optimal balance between acquiring new knowledge and preserving generalization. Importantly, this also suggests a promising direction for future work by combining complementary strategies rather than relying on a single one.
\end{itemize}

\setlength{\tabcolsep}{2mm}
\begin{table}[t]
\centering
\footnotesize
\caption{Ablation results averaged over 8 datasets (base-to-base/novel generalization) and across all domains (leave-one-domain-out generalization).}
\resizebox{0.9\textwidth}{!}{ 
\begin{tabular}{c|>{\centering\arraybackslash}p{1.5cm}>{\centering\arraybackslash}p{1.5cm}>{\centering\arraybackslash}p{1.5cm}|cc}
\toprule
\multirow{2}{*}{Method} & \multicolumn{3}{c|}{\textbf{Base to Base/Novel Generalization}} & \multicolumn{2}{c}{\textbf{Leave-One-Domain-Out Generalization}} \\\cmidrule(lr){2-6}
& Base & Novel & HM & Office-Home & DomainNet \\
\midrule
PromptFL            & 64.06 & 71.48 & 66.98 & 75.17 & 84.24 \\
+ AttnIso$_{\text{HA}}$       & 62.19 & \textbf{72.27} & 67.04 & 77.26 & 84.87 \\
+ AttnIso$_{\text{RE}}$       & 64.22 & 71.51 & 67.76 & 75.67 & 84.36 \\
+ AttnIso$_{\text{HA+RE}}$        & 64.11 & 72.66 & 67.87 & 77.83 & 85.01 \\
+ CSCR      & \textbf{73.34} & 64.89 & 67.24 & 78.59 & 85.11 \\
+ AttnIso$_{\text{HA}}$ + CSCR & 69.32 & 70.78 & 70.06 & 79.13 & 85.46 \\
+ AttnIso$_{\text{RE}}$ + CSCR & 70.56 & 68.11 & 69.65 & 78.42 & 85.23 \\
+ AttnIso$_{\text{HA+RE}}$ + CSCR & 72.32 & 69.47 & \textbf{70.72} & \textbf{80.57} & \textbf{85.76} \\
\bottomrule
\end{tabular}}
\label{tab3}
\end{table}

\begin{figure}[ht]
\centering
\includegraphics[width=1.0\linewidth]{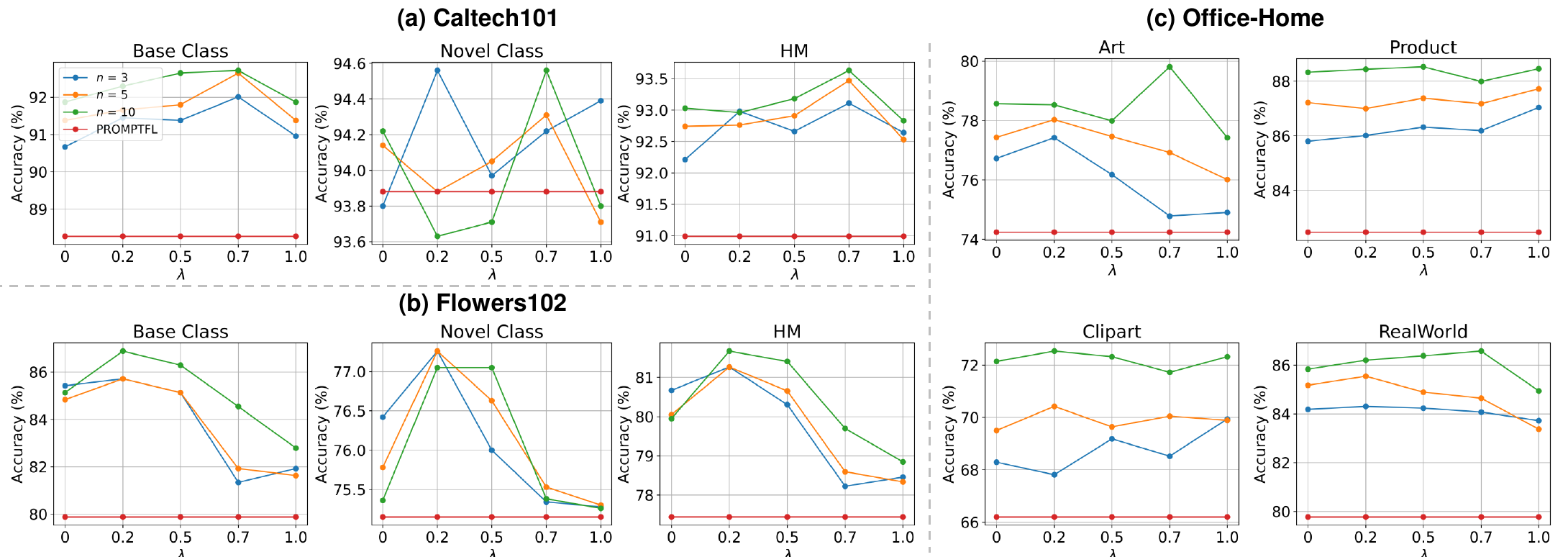}
\caption{The impact of hyperparameters on performance is evaluated by varying $\lambda$ from $\{0, 0.2, 0.5, 0.7, 1.0\}$, the number of prototypes per class $n$ from $\{3, 5, 10\}$ on three datasets.
} 
\label{fig}
\end{figure}

\subsection{Sensitivity Analysis}
This section evaluates the robustness of \methodname{} across varying hyperparameter settings on the DTD, Flowers102 and Office-Home datasets. Specifically, we investigate the effects of hyperparameters $n$ (the number of prototypes generated per class on the client side) and $\lambda$ (the weighting factor in the AttnIso module), evaluated over $\{3, 5, 10\}$ and $\{0, 0.2, 0.5, 0.7, 1.0\}$, respectively. As depicted in Figure \ref{fig}, \methodname{} consistently achieves higher overall HM scores than its baselines in all cases of the base-to-base/novel generalization task, showcasing its superior balance between retaining transferable knowledge and learning task-specific knowledge. On the one hand, \methodname{} demonstrates consistent improvements over baselines on base classes, which benefits from the CSCR module’s capacity to alleviate data heterogeneity. On the other hand \methodname{} exhibits competitive performance on novel-class generalization, frequently outperforming other methods, which is attributed to the attention masking mechanism of the AttnIso module. Furthermore, in the Leave-One-Domain-Out Generalization task, similar conclusions are drawn, as \methodname{} outperforms the baseline across multiple domains. 

\subsection{Performance Comparison across Different Local Training Epoch Settings}
\begin{table*}[t]
\centering
\caption{Performance comparison between \methodname{} and baselines on DTD and Flowers102 with local training epochs $\in \{1, 5, 20\}$ in the base-to-base/novel generalization setting.}
\renewcommand{\arraystretch}{1.2}
\resizebox{1.0\textwidth}{!}{
\begin{tabular}{lccc|lccc|lccc}
\toprule
\multicolumn{4}{c|}{\textbf{(a) DTD (Epochs=1)}} & 
\multicolumn{4}{c|}{\textbf{(b) DTD (Epochs=5)}} & 
\multicolumn{4}{c}{\textbf{(c) DTD (Epochs=20)}} \\
\hline
\textbf{Method} & \textbf{Base} & \textbf{Novel} & \textbf{HM} & 
\textbf{Method} & \textbf{Base} & \textbf{Novel} & \textbf{HM} & 
\textbf{Method} & \textbf{Base} & \textbf{Novel} & \textbf{HM} \\
\midrule
CLIP & 53.35 & 59.05 & 56.06 & CLIP & 53.35 & 59.05 & 56.06 & CLIP & 53.35 & 59.05 & 56.06 \\
PromptFL & 50.34 & 51.69 & 51.01 & PromptFL & 55.55 & 56.76 & 56.15 & PromptFL & 63.31 & 51.08 & 56.54 \\
RPO & 52.77 & \textbf{59.42} & 55.90 & RPO & 53.24 & \textbf{59.78} & 56.32 & RPO & 53.93 & \textbf{60.26} & 56.92 \\
FedTPG & 40.27 & 36.47 & 38.28 & FedTPG & 49.88 & 47.22 & 48.51 & FedTPG & 56.25 & 56.28 & 56.26 \\
TCP & 53.24 & 53.38 & 53.31 & TCP & 57.63 & 56.40 & 57.01 & TCP & 63.07 & 59.54 & 61.25 \\
PromptFolio & 52.54 & 53.50 & 53.02 & PromptFolio & 58.68 & 56.28 & 57.45 & PromptFolio & 58.68 & 57.60 & 58.13 \\
DP-FPL & 43.75 & 41.54 & 42.61 & DP-FPL & 59.25 & 50.12 & 54.30 & DP-FPL & 64.12 & 47.82 & 54.78 \\\hline
\methodname{}$_{P}$ & \textbf{65.97} & 51.20 & \textbf{57.65} & \methodname{}$_{P}$ & \textbf{68.51} & 52.65 & 59.54 & \methodname{}$_{P}$ & \textbf{70.13} & 53.26 & 60.54 \\
\methodname{}$_{T}$ & 63.07 & 51.81 & 56.89 & \methodname{}$_{T}$ & 65.85 & 55.07 & \textbf{59.98} & \methodname{}$_{T}$ & 69.67 & 54.83 & \textbf{61.36} \\
\midrule
\multicolumn{4}{c|}{\textbf{(d) Flowers102 (Epochs=1)}} & 
\multicolumn{4}{c|}{\textbf{(e) Flowers102 (Epochs=5)}} & 
\multicolumn{4}{c}{\textbf{(f) Flowers102 (Epochs=20)}} \\
\hline
\textbf{Method} & \textbf{Base} & \textbf{Novel} & \textbf{HM} & 
\textbf{Method} & \textbf{Base} & \textbf{Novel} & \textbf{HM} & 
\textbf{Method} & \textbf{Base} & \textbf{Novel} & \textbf{HM} \\
\midrule
CLIP & 62.39 & 78.10 & 69.36 & CLIP & 62.39 & 78.10 & 69.36 & CLIP & 62.39 & 78.10 & 69.36 \\
PromptFL & 70.84 & 74.10 & 72.43 & PromptFL & 78.42 & 76.00 & 77.19 & PromptFL & 80.46 & 73.68 & 76.92 \\
RPO & 69.97 & 76.21 & 72.95 & RPO & 71.72 & 76.84 & 74.19 & RPO & 74.05 & 77.05 & 75.52 \\
FedTPG & 68.51 & 77.05 & 72.53 & FedTPG & 73.76 & 77.05 & 75.37 & FedTPG & 74.34 & 77.05 & 75.67 \\
TCP & 75.51 & 78.10 & 76.78 & TCP & 76.09 & 79.15 & 77.59 & TCP & 79.88 & 78.31 & 79.09 \\
PromptFolio & 73.76 & 76.42 & 75.06 & PromptFolio & 79.59 & 75.57 & 77.53 & PromptFolio & 83.09 & 74.52 & 78.57 \\
DP-FPL & 69.38 & \textbf{78.31} & 73.58 & DP-FPL & 73.76 & \textbf{79.57} & 76.55 & DP-FPL & 75.80 & 75.78 & 75.79 \\\hline
\methodname{}$_{P}$ & 82.50 & 75.57 & 78.89 & \methodname{}$_{P}$ & 84.83 & 78.31 & 81.44 & \methodname{}$_{P}$ & 85.13 & 77.47 & 81.12 \\
\methodname{}$_{T}$ & 87.11 & 77.83 & \textbf{82.20} & \methodname{}$_{T}$ & \textbf{86.58} & 78.94 & \textbf{82.59} & \methodname{}$_{T}$ & \textbf{87.75} & \textbf{78.73} & \textbf{83.00} \\
\bottomrule
\end{tabular}
}
\label{tab7}
\end{table*}

To assess the robustness of the proposed \methodname{} under different local training efforts, we evaluate its performance with local epochs $\in {1, 5, 20}$, a data heterogeneity level of $\beta=0.5$, and 10 participating clients. As shown in Table~\ref{tab7}, both \methodname{}$_{P}$ and \methodname{}$_{T}$ consistently outperform baselines across all datasets and settings. Notably, \methodname{}$_{T}$ achieves the highest harmonic mean (HM) in most cases, indicating better generalization across base and novel classes. On the Flowers102 dataset, it maintains strong performance with increasing local epochs, reflecting robustness against overfitting. Overall, both variants remain stable and effective despite changing local training dynamics.

\subsection{Effectiveness of Global Prompt Refinement}
\begin{figure}[t]
\centering
\includegraphics[width=1\linewidth]{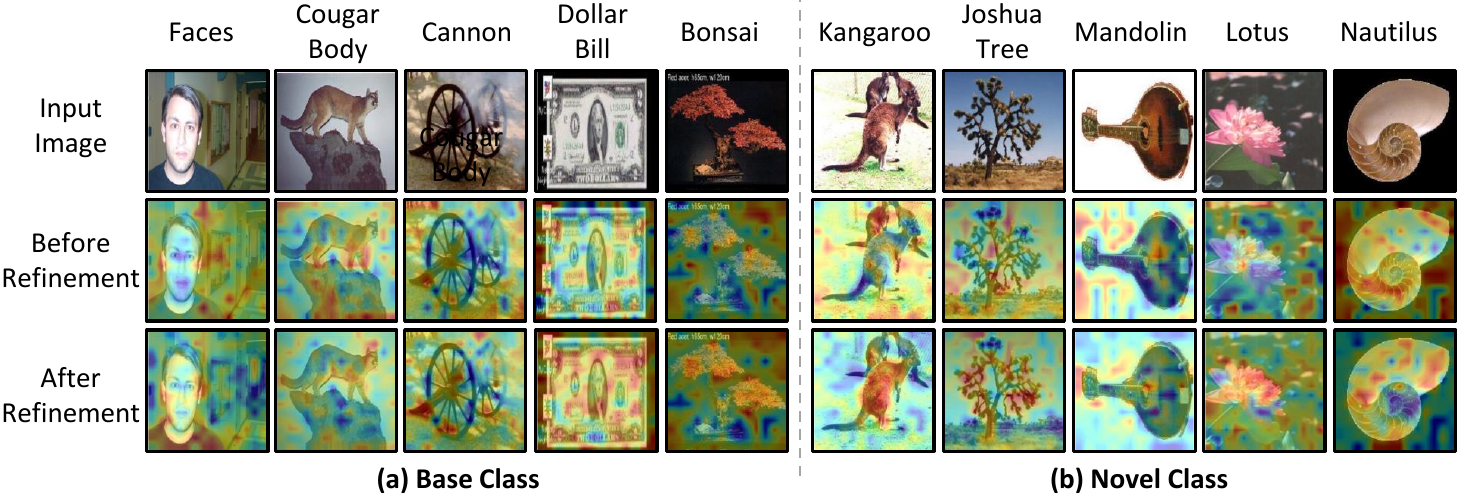}
\caption{Visualization of similarity heatmaps between text features and image patches before and after prompt refinement on Caltech101. In the similarity maps, red and yellow indicate higher similarity between image patches and the text features, while blue represents lower similarity.
} 
\label{fig9}
\end{figure}
This section aims to evaluate the effectiveness of the global prompt refinement method. Specifically, we visualize the similarity maps between text features and image patches before and after refinement, as shown in Figure \ref{fig9}. 
Intuitively, the similarity responses between visual targets and textual semantics are often scattered and misaligned before refinement. For example, in the ``Dollar Bill'' and ``Bonsai'' samples, the high similarity responses in the second row primarily concentrate on background regions near the image boundaries. In contrast, the responses become significantly more concentrated on semantically relevant regions, such as the face on the bill or the foliage of the bonsai after refinement. Notably, this advantage is also evident in novel classes. For instance, the refined prompts enable the model to better localize foreground objects while suppressing irrelevant background activations in ``Kangaroo'' and ``Nautilus'' samples. Moreover, it becomes tightly aligned with the spiral structure of the shell in the ``Nautilus'', indicating enhanced semantic grounding.
\begin{figure}[htbp]
  \begin{minipage}[t]{0.48\linewidth}
    \centering
\includegraphics[width=1.\linewidth]{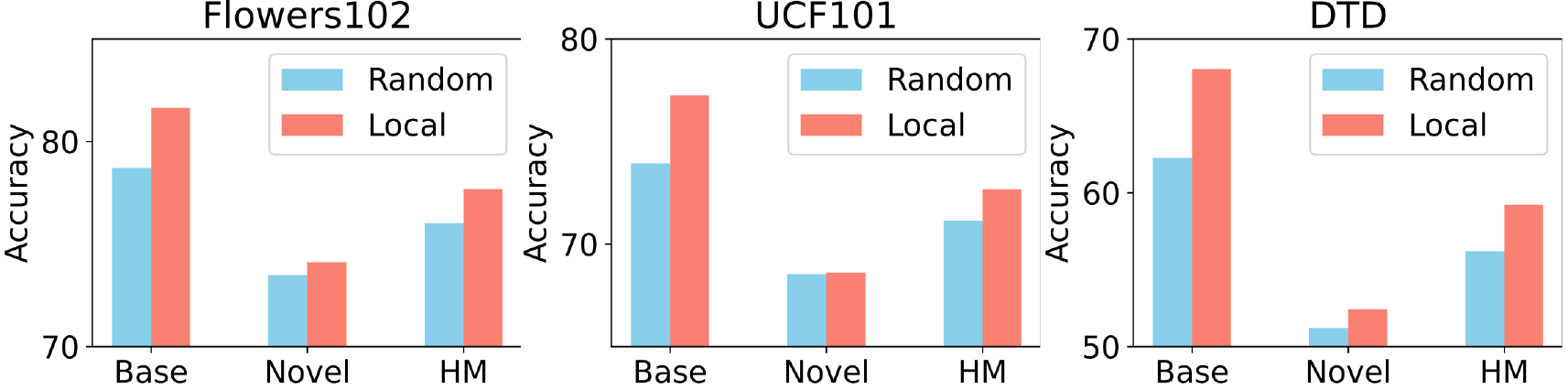}
    \caption{Comparison between Random and Local Prompt-Tuning for Initialization.}
    \label{fig5}
  \end{minipage}%
  \hfill
  \begin{minipage}[t]{0.48\linewidth}
    \centering
\includegraphics[width=1\linewidth]{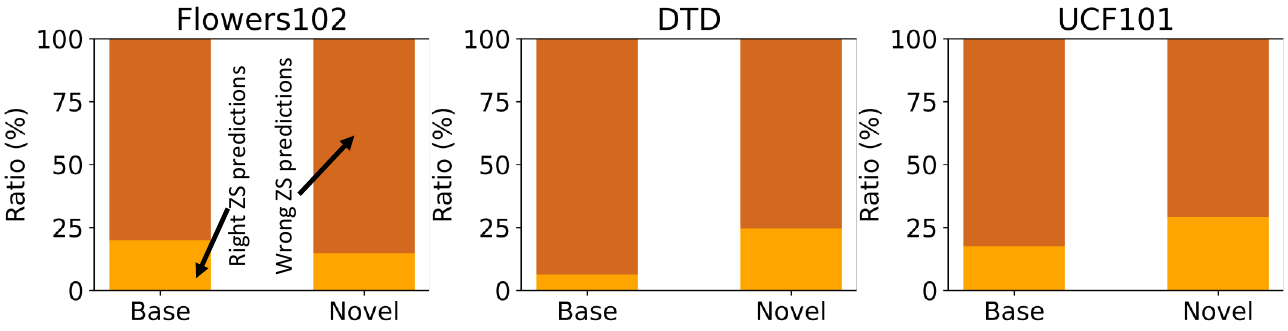}
    \caption{Failure case analysis. Most \methodname{} errors overlap with zero-shot CLIP.}
    \label{fig6}
  \end{minipage}
\vspace{-0.5cm}
\end{figure}

\subsection{Evaluation of Local Prompt Tuning as Initialization}
This section evaluates the performance of random initialization and local prompt tuning (LPT) on Flowers102, UCF101 and DTD datasets. As shown in Figure \ref{fig5}, LPT yields superior performance compared to random initialization on all cases. This can be attributed to the local training that can be more aligned with the target data, allowing the model to better capture domain-specific features. In particular, LPT outperforms the random initialization by 6\% on the DTD dataset for base classes, which highlights the method's ability to adapt more efficiently to the nuances of the texture.
\subsection{Failure Case Analysis} \label{sec5.7}
This section reports the percentage of cases where the zero-shot CLIP model also failed in instances where the \methodname{} model made incorrect predictions (Figure \ref{fig6}). Clearly, there is a high overlap in the failure cases of both models. This reflects that the imprecision of zero-shot generalized knowledge hinders the model's generalization, which also provides insights for future work to address the challenges posed by the lack of domain-specific adaptation in generalization models.

\section{Conclusions and Future Work}
In this paper, we balance the model capability to learn task-specific knowledge for downstream tasks with its capacity to retain generalized knowledge. Specifically, we propose \methodname{} for one-shot FPL, which introduces an attention masking mechanism to limit the interaction between the learnable prompt tokens and the original text tokens. Furthermore, it learns a unified prompt through global prompt refinement, thereby mitigating the data heterogeneity issue. Experimental results demonstrate that \methodname{} significantly improves the performance of global prompts across diverse visual tasks.

Despite the promising results of \methodname{}, there remains room for improvement. The communication overhead from prototype transmission may become a bottleneck in large-scale federated systems with limited bandwidth, which we plan to address in future work. And both the multi-label \cite{kou2025labeldistributionlearningbiased,ijcai2024p478}, multi-modal \cite{meng2025causal,zhengwang_tnnls2025,wang2023multi,cheng2023voice} problems and real-world medical imaging tasks \cite{feng2023unsupervised,feng2025neighbor,dang2025deep} become considerably more challenging in federated settings. In addition, several issues discussed in Section~\ref{sec5.7} remain open and deserve further investigation.

\section*{Acknowledgments}
This work is supported in part by the Key Research and Development Program of Shandong Province (Grant No. 2024TSGC0667); the Ministry of Education, Singapore, under its Academic Research Fund Tier 1 (RG101/24); the RIE2025 Industry Alignment Fund – Industry Collaboration Projects (IAF-ICP) (Award I2301E0026), administered by A*STAR, as well as supported by Alibaba Group and NTU Singapore through Alibaba-NTU Global e-Sustainability CorpLab (ANGEL).

\small
\bibliographystyle{unsrt}
\bibliography{NIPS2025references}

\appendix
\newpage
\section{Appendix}
\subsection{Datasets}
To conduct our evaluation, we selected 10 diverse visual classification datasets as benchmarks. Table \ref{tab4} provides a detailed overview, including the number of classes, the size of training and testing sets, the number of domains, and the corresponding federated settings.

\begin{figure}[h]
\centering
\includegraphics[width=1\linewidth]{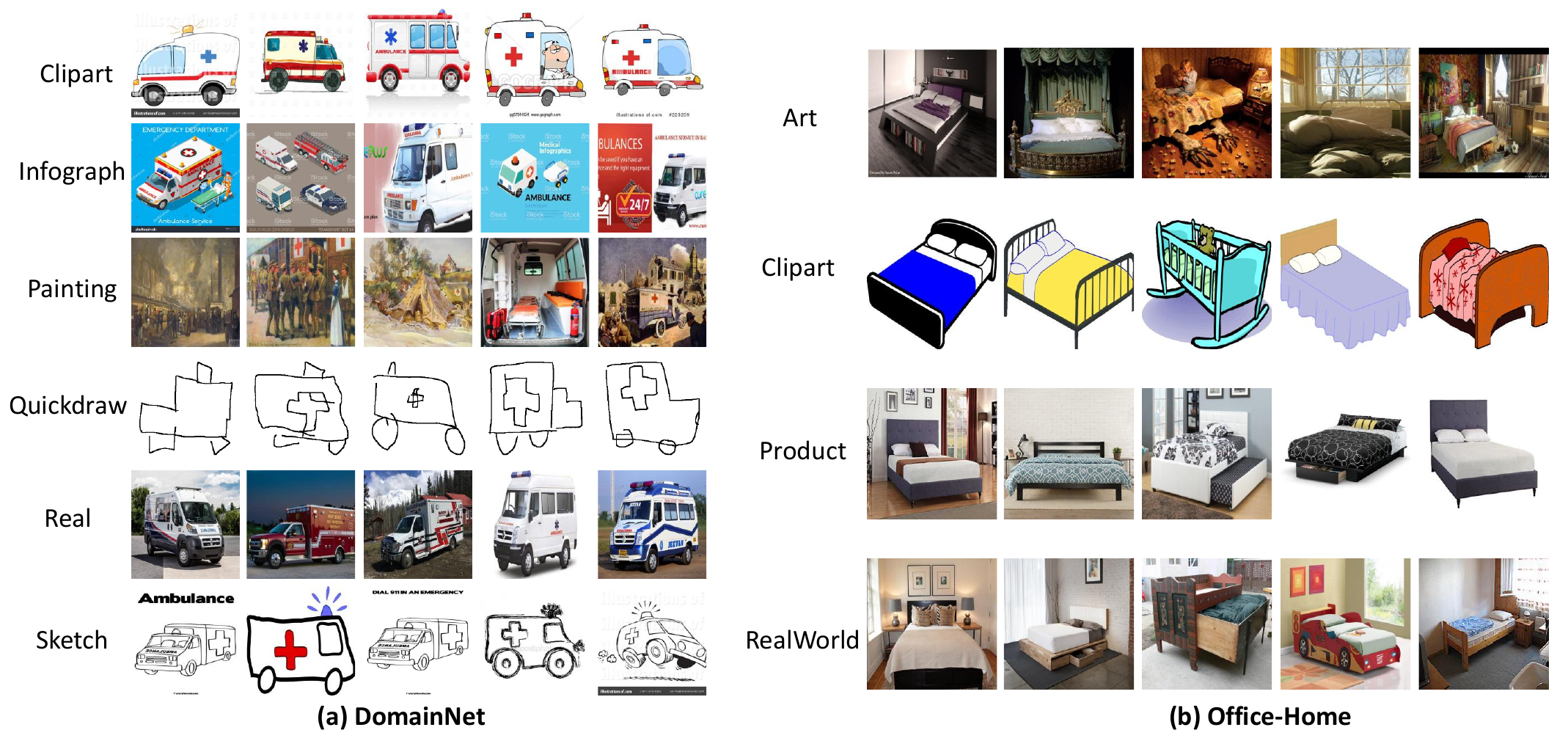}
\caption{Illustrative examples of class instances across domains: “Ambulance” from DomainNet and “Bed” from Office-Home.
} 
\label{fig7}
\end{figure}

For datasets with multiple domains, we adopt the commonly used Office-Home benchmark, which includes four domains: Art, Clipart, Product, and RealWorld. In addition, we utilize DomainNet, a large-scale dataset spanning six domains: Clipart, Infograph, Painting, Quickdraw, Real, and Sketch. We focus on training with a selected subset of 20 classes from each dataset. Figure \ref{fig7} illustrates representative raw instances from the two multi-domain datasets.

\begin{table}[h]
\centering
\caption{Statistics and federated settings of the datasets used in our experiments.}
\label{tab:dataset_stats}
\renewcommand{\arraystretch}{1.1}
\begin{tabular}{l|c|c|c|c|c|c|c}
\toprule
\multirow{2}{*}{\textbf{Dataset}} & \multirow{2}{*}{\textbf{Classes}} & \multirow{2}{*}{\textbf{Train}} & \multicolumn{2}{c|}{\textbf{Test}} & \multirow{2}{*}{\textbf{Domains}} & \multicolumn{2}{c}{\textbf{Federated Settings}} \\
\cmidrule(lr){4-5} \cmidrule(lr){7-8}
& & & \textbf{Base} & \textbf{Novel} & & \textbf{Clients} & \textbf{Heterogeneity} \\
\midrule
CIFAR10        & 10   & 25,000  & 5,000  & 5,000  & 1 & 5/10/20    & 0.1/0.5 \\
OxfordPets     & 37   & 1,785   & 1,782  & 1,887  & 1 & 5/10/20   & 0.1/0.5 \\
Caltech101     & 100  & 3,310   & 1,416  & 1,178  & 1 & 5/10/20   & 0.1/0.5 \\
DTD            & 47   & 1,440   & 864    & 828    & 1 & 5/10/20   & 0.1/0.5 \\
FGVCAircraft   & 100  & 3,333   & 1,667  & 1,666  & 1 & 5/10/20   & 0.1/0.5 \\
Flowers102     & 102  & 2,807   & 343  & 475  & 1 & 5/10/20   & 0.1/0.5 \\
StanfordCars   & 196  & 4,052   & 4,002  & 3,998  & 1 & 5/10/20   & 0.1/0.5 \\
UCF101         & 101  & 3,926   & 1,895  & 1,888  & 1 & 5/10/20   & 0.1/0.5 \\
\midrule
DomainNet      & 20   & \multicolumn{3}{c|}{30,777}   & 6 & 9/15   & -- \\
Office-Home    & 65   & \multicolumn{3}{c|}{15,588}    & 4 & 15/25   & -- \\
\bottomrule
\end{tabular}
\label{tab4}
\end{table}

\subsection{Experimental Results}
\subsubsection{Base-to-Base/Novel Generalization under High-Level Data Heterogeneity}
\begin{table*}[h]
\centering
\caption{Performance comparison between \methodname{} and baselines on 8 datasets with $\beta=0.1$ in the base-to-base/novel generalization setting. The best performance is indicated in bold.}
\renewcommand{\arraystretch}{1.2} 
\resizebox{1.0\textwidth}{!}{
\begin{tabular}{lccc|lccc|lccc}
\toprule
\multicolumn{4}{c|}{\textbf{(a) Average over 8 datasets}} & 
\multicolumn{4}{c|}{\textbf{(b) CIFAR10}} & 
\multicolumn{4}{c}{\textbf{(c) OxfordPets}} \\\hline
\textbf{Method} & \textbf{Base} & \textbf{Novel} & \textbf{HM} & 
\textbf{Method} & \textbf{Base} & \textbf{Novel} & \textbf{HM} & 
\textbf{Method} & \textbf{Base} & \textbf{Novel} & \textbf{HM} \\
\midrule
CLIP        & 65.79 & 74.32 & 69.70 & CLIP      & 95.30 & 95.50 & 95.39 & CLIP        & 81.53 & 95.33 & 87.89 \\
PromptFL    & 69.15 & 72.50 & 70.71 & PromptFL    & 94.72 & 95.68 & 95.19 & PromptFL    & 89.33 & 95.91 & 92.51 \\
RPO & 67.81 & 73.66 & 70.57 & RPO & 95.10 & 94.76 & 94.92 & RPO & 86.19 & 95.54 & 90.63 \\
FedTPG      & 67.06 & 71.71 & 69.20 & FedTPG      & 94.30 & 96.64 & 95.45 & FedTPG      & 86.58 & 94.70 & 90.46 \\
TCP         & 70.30 & 74.19 & 72.13 & TCP         & 95.30 & 95.66 & 95.47 & TCP         & 90.01 & \textbf{96.23} & 93.02 \\
PromptFolio & 69.49 & 73.44 & 71.33 & PromptFolio & 95.20 & 96.74 & 95.96 & PromptFolio & 90.90 & 95.17 & 92.99 \\
DP-FPL & 71.90 & 71.75 & 69.88 & DP-FPL & 94.24 & 95.42 & 94.82 & DP-FPL & 91.30 & 95.12 & \textbf{93.17} \\
FedDISC     & 52.27 & 54.40 & 50.47 & FedDISC     & 92.47 & 91.37 & 91.91 & FedDISC     & 81.48 & 66.74 & 73.38 \\
FedELMY      & 68.05 & 47.29 & 53.41 & FedELMY      & 89.84 & 28.90 & 43.73 & FedELMY      & 86.70 & 82.77 & 84.69 \\\hline
\methodname{}$_{P}$      & \textbf{73.87} & 72.84 & 73.18 & \methodname{}$_{P}$      & 95.88 & \textbf{96.84} & \textbf{96.35} & \methodname{}$_{P}$      & 90.57 & 95.23 & 92.84 \\
\methodname{}$_{T}$      & 73.62 & \textbf{74.48} & \textbf{73.97} & \methodname{}$_{T}$      & \textbf{96.03} & 96.24 & 96.13 & \methodname{}$_{T}$      & \textbf{91.86} & 94.38 & 93.10 \\
\bottomrule
\multicolumn{4}{c|}{\textbf{(d) Caltech101}} & 
\multicolumn{4}{c|}{\textbf{(e) DTD}} & 
\multicolumn{4}{c}{\textbf{(f) FGVCAircraft}} \\\hline
\textbf{Method} & \textbf{Base} & \textbf{Novel} & \textbf{HM} & 
\textbf{Method} & \textbf{Base} & \textbf{Novel} & \textbf{HM} & 
\textbf{Method} & \textbf{Base} & \textbf{Novel} & \textbf{HM} \\
\midrule
CLIP        & 82.27 & 94.05 & 87.77 & CLIP        & 53.35 & 59.05 & 56.06 & CLIP        & 26.51 & 31.21 & 28.67 \\
PromptFL    & 87.07 & 93.03 & 89.95 & PromptFL    & 56.36 & 52.53 & 54.38 & PromptFL    & 26.03 & 30.61 & 28.13 \\
RPO & 84.95 & 93.80 & 89.16 & RPO & 53.12 & \textbf{60.02} & 56.36 & RPO & 26.51 & 30.37 & 28.31 \\
FedTPG      & 85.38 & 93.37 & 89.20 & FedTPG      & 53.35 & 50.00 & 51.62 & FedTPG      & 21.17 & 26.35 & 23.48 \\
TCP         & 88.41 & \textbf{94.16} & 91.19 & TCP         & 58.10 & 57.60 & 57.85 & TCP         & 28.37 & 32.23 & 30.18 \\
PromptFolio & 82.41 & 92.61 & 87.21 & PromptFolio & 53.58 & 55.55 & 54.55 & PromptFolio & 28.13 & \textbf{34.15} & \textbf{30.85} \\
DP-FPL & 87.57 & 93.54 & 90.46 & DP-FPL & 51.85 & 45.65 & 48.55 & DP-FPL & 26.63 & 29.35 & 27.92 \\
FedDISC     & 22.37 & 81.39 & 35.09 & FedDISC     & 65.23 & 41.23 & 49.87 & FedDISC     & 18.49 & 17.61 & 18.03 \\
FedELMY      & 85.59 & 90.40 & 87.93 & FedELMY      & 40.27 & 11.95 & 18.43 & FedELMY      & 23.45 & 5.46 & 8.86 \\\hline
\methodname{}$_{P}$      & \textbf{91.24} & 93.88 & \textbf{92.54} & \methodname{}$_{P}$      & \textbf{67.59} & 52.89 & 59.34 & \methodname{}$_{P}$      & \textbf{29.69} & 30.81 & 30.23 \\
\methodname{}$_{T}$      & 90.60 & 94.05 & 92.30 & \methodname{}$_{T}$      & 65.62 & 57.79 & \textbf{61.45} & \methodname{}$_{T}$      & 29.39 & 29.96 & 29.67 \\
\bottomrule
\multicolumn{4}{c|}{\textbf{(g) Flowers102}} & 
\multicolumn{4}{c|}{\textbf{(h) StanfordCars}} & 
\multicolumn{4}{c}{\textbf{(i) UCF101}} \\\hline
\textbf{Method} & \textbf{Base} & \textbf{Novel} & \textbf{HM} & 
\textbf{Method} & \textbf{Base} & \textbf{Novel} & \textbf{HM} & 
\textbf{Method} & \textbf{Base} & \textbf{Novel} & \textbf{HM} \\
\midrule
CLIP        & 62.39 & 78.10 & 69.36 & CLIP        & 56.87 & 69.40 & 62.51 & CLIP        & 68.12 & 71.92 & 69.97 \\
PromptFL    & 72.88 & 75.78 & 74.30 & PromptFL    & 57.31 & 67.58 & 62.02 & PromptFL    & 69.55 & 68.90 & 69.22 \\
RPO & 72.01 & 76.63 & 74.24 & RPO & 57.64 & 66.73 & 61.85 & RPO & 67.01 & 71.45 & 69.16 \\
FedTPG      & 70.84 & 77.26 & 73.91 & FedTPG      & 56.89 & \textbf{69.55} & 62.59 & FedTPG      & 68.02 & 65.83 & 66.91 \\
TCP         & 73.61 & 77.94 & 75.71 & TCP         & 57.79 & 69.28 & 63.01 & TCP         & 70.88 & 70.43 & 70.65 \\
PromptFolio & 75.51 & 76.42 & 75.96 & PromptFolio & 58.97 & 69.05 & 63.61 & PromptFolio & 71.29 & 67.84 & 69.52 \\
DP-FPL & 68.97 & 71.43 & 70.08 & DP-FPL & 53.99 & 63.65 & 58.43 & DP-FPL & 73.61 & 69.80 & 71.66 \\
FedDISC     & 25.44 & 32.47 & 28.52 & FedDISC     & 42.59 & 51.23 & 46.51 & FedDISC     & 70.12 & 53.23 & 60.51 \\
FedELMY      & \textbf{81.63} & 57.05 & 67.16 & FedELMY      & 60.26 & 49.94 & 54.62 & FedELMY      & \textbf{76.72} & 51.90 & 61.92 \\\hline
\methodname{}$_{P}$      & 79.30 & 77.26 & 78.26 & \methodname{}$_{P}$      & \textbf{60.81} & 66.98 & 63.75 & \methodname{}$_{P}$      & 75.88 & 68.85 & 72.19 \\
\methodname{}$_{T}$      & 80.17 & \textbf{79.15} & \textbf{79.65} & \methodname{}$_{T}$      & 60.14 & 68.65 & \textbf{64.12} & \methodname{}$_{T}$      & 75.19 & \textbf{75.63} & \textbf{75.41} \\
\bottomrule
\end{tabular}
}
\label{tab5}
\end{table*}
To more comprehensively evaluate the effectiveness of the methods, we test all approaches on eight datasets under a higher level of heterogeneity ($\beta = 0.1$), focusing on their performance in the base-to-base/novel generalization setting. As presented in Table \ref{tab5}, our proposed method \methodname{} obtains the highest average performance on the eight datasets. Specifically, \methodname{}$_P$ and \methodname{}$_T$ achieve the top two performances on the base classes, ranking first and second, respectively. In terms of novel class generalization, \methodname{}$_T$ outperforms all baselines, demonstrating its superior capability to adapt to unseen categories. Furthermore, \methodname{}$_T$ also achieves the highest overall performance in terms of harmonic mean (HM), indicating a strong balance between knowledge retention and generalization. These results highlight the robustness and adaptability of the proposed method under highly heterogeneous federated settings.

\subsubsection{Leave-One-Domain-Out Generalization with Increased Client Diversity}

\begin{table*}[t]
\centering
\caption{Performance comparison between \methodname{} and baselines on the Office-Home and DomainNet datasets under the leave-one-domain-out generalization setting, where the data from each domain is randomly partitioned among 5 clients.}
\renewcommand{\arraystretch}{1.2}
\resizebox{\textwidth}{!}{
\begin{tabular}{l|ccccc|cccccccc}
\toprule
\multirow{2}{*}{\textbf{Method}} & 
\multicolumn{5}{c|}{\textbf{Office-Home}}  & 
\multicolumn{7}{c}{\textbf{DomainNet}} \\
\cmidrule(lr){2-6} \cmidrule(lr){7-13}
& Art & Clipart & Product & RealWorld & Average & 
Clipart & Infograph & Painting & Quickdraw & Real & Sketch & Average\\
\midrule
CLIP      & 74.57 & 62.33 & 80.17 & 79.38 & 74.08 & 92.87 & 77.02 & 85.35 & 43.06 & 98.61 & 94.22 & 81.85 \\
PromptFL      & 73.71 & 64.49 & 82.81 & 80.39 & 75.35 & 94.93 & 80.09 & \textbf{87.94} & 52.68 & 99.00 & 95.63 & 85.04 \\
RPO     & 73.83 & 62.13 & 79.95 & 79.36 & 73.81 & 93.66 & 77.31 & 84.65 & 44.46 & 98.83 & 93.87 & 82.13 \\
FedTPG       & 74.90 & 65.54 & 82.85 & 80.37 & 75.91 & 94.35 & 79.76 & 87.40 & 48.05 & 98.92 & \textbf{95.89} & 84.06 \\
TCP       & 75.77 & 65.10 & 83.10 & 80.99 & 76.24 & 93.92 & 79.72 & 87.67 & 46.82 & 98.96 & 95.00 & 83.68 \\
PromptFolio     & 75.81 & 65.36 & 83.53 & 80.39 & 76.27 & \textbf{95.51} & \textbf{81.00} & 89.33 & 48.82 & 98.99 & 95.20 & 84.80 \\
DP-FPL   & 74.82 & 63.55 & 82.85 & 78.79 & 75.00 & 94.77 & 80.20 & 89.00 & 48.35 & 98.78 & 94.82 & 84.32 \\
FedDISC     & 71.19 & 62.48 & 81.31 & 78.35 & 73.48 & 94.23 & 80.25 & 83.85 & 44.21 & 98.39 & 92.49 & 82.23 \\
FedELMY   & 75.45 & 64.29 & 82.36 & 77.58 & 74.92 & 94.15 & 75.48 & 86.89 & 47.39 & 98.72 & 95.13 & 82.96 \\\hline
\methodname{}$_P$     & \textbf{78.16} & \textbf{72.76} & 89.09 & 83.68 & 80.92 & 94.45 & 80.60 & 87.50 & \textbf{58.68} & \textbf{99.10} & 94.91 & 85.87 \\
\methodname{}$_T$      & 77.66 & 71.95 & \textbf{89.47} & \textbf{85.93} & \textbf{81.25} & 95.19 & 80.49 & 87.23 & 57.96 & 99.07 & 95.37 & \textbf{85.88} \\
\bottomrule
\end{tabular}
}
\label{tab6}
\end{table*}
To validate the effectiveness of the proposed \methodname{} under more challenging scenarios, we conduct experiments in a leave-one-domain-out generalization setting \cite{} using the Office-Home and DomainNet datasets. In this setting, the data from each domain is randomly partitioned among 5 clients, and for each evaluation, one domain is held out entirely for testing while the remaining domains are used for federated training. As shown in Table~\ref{tab6}, \methodname{}$_P$ and \methodname{}$_T$ consistently outperform existing baselines across both datasets. For the Office-Home, \methodname{}$_T$ achieves the highest average accuracy, with \methodname{}$_P$ closely following. Notably, \methodname{}$_T$ yields the best performance on the ``RealWorld" and ``Product" domains, while \methodname{}$_P$ leads on ``Art" and ``Clipart". For the DomainNet, both variants of \methodname{} also demonstrate strong cross-domain generalization, achieving top results in multiple domains. These results highlight the robustness of our method in handling both domain shift and non-IID data across clients. In summary, the proposed \methodname{} exhibits superior generalization performance under leave-one-domain-out conditions, demonstrating its effectiveness in federated scenarios with high domain and client-level variability.

\subsubsection{Performance Comparison across Different Local Training Epoch Settings}

\begin{table*}[h]
\centering
\caption{Performance comparison between \methodname{} and baselines on StanfordCars and UCF101 with local training epochs $\in \{1, 5, 20\}$ in the base-to-base/novel generalization setting.}
\renewcommand{\arraystretch}{1.2}
\resizebox{1.0\textwidth}{!}{
\begin{tabular}{lccc|lccc|lccc}
\toprule
\multicolumn{4}{c|}{\textbf{(g) StanfordCars (Epochs=1)}} & 
\multicolumn{4}{c|}{\textbf{(h) StanfordCars (Epochs=5)}} & 
\multicolumn{4}{c}{\textbf{(i) StanfordCars (Epochs=20)}} \\\hline
\textbf{Method} & \textbf{Base} & \textbf{Novel} & \textbf{HM} & 
\textbf{Method} & \textbf{Base} & \textbf{Novel} & \textbf{HM} & 
\textbf{Method} & \textbf{Base} & \textbf{Novel} & \textbf{HM} \\
\midrule
CLIP        & 56.87 & 69.40 & 62.51 & CLIP        & 56.87 & 69.40 & 62.51 & CLIP        & 56.87 & 69.40 & 62.51 \\
PromptFL    & 58.87 & 68.40 & 63.27 & PromptFL    & 60.39 & 67.35 & 63.68 & PromptFL    & 59.74 & 67.83 & 63.52 \\
RPO & 56.39 & 69.03 & 62.07 & RPO & 57.12 & 69.70 & 62.79 & RPO & 57.42 & \textbf{70.06} & 63.11 \\
FedTPG      & 57.09 & 69.73 & 62.78 & FedTPG      & 58.14 & 69.65 & 63.38 & FedTPG      & 58.69 & 69.25 & 63.54 \\
TCP         & 58.04 & 69.40 & 63.22 & TCP         & 58.72 & 69.91 & 63.82 & TCP         & 59.29 & 68.68 & 63.64 \\
PromptFolio & 58.69 & \textbf{69.83} & 63.78 & PromptFolio & 60.04 & 69.78 & 64.54 & PromptFolio & 61.46 & 69.10 & 65.06 \\
DP-FPL & 49.30 & 62.85 & 55.25 & DP-FPL & 59.02 & 68.33 & 63.33 & DP-FPL & 59.24 & 64.35 & 61.69 \\
\hline
\methodname{}$_{P}$      & 60.61 & 67.58 & 63.91 & \methodname{}$_{P}$      & 62.09 & 67.05 & 64.48 & \methodname{}$_{P}$      & \textbf{64.46} & 66.38 & 65.41 \\
\methodname{}$_{T}$      & \textbf{61.54} & 67.40 & 64.33 & \methodname{}$_{T}$      & 60.99 & 69.45 & \textbf{64.94} & \methodname{}$_{T}$      & 63.01 & 69.00 & \textbf{65.87} \\
\bottomrule
\multicolumn{4}{c|}{\textbf{(j) UCF101 (Epochs=1)}} & 
\multicolumn{4}{c|}{\textbf{(k) UCF101 (Epochs=5)}} & 
\multicolumn{4}{c}{\textbf{(l) UCF101 (Epochs=20)}} \\\hline
\textbf{Method} & \textbf{Base} & \textbf{Novel} & \textbf{HM} & 
\textbf{Method} & \textbf{Base} & \textbf{Novel} & \textbf{HM} & 
\textbf{Method} & \textbf{Base} & \textbf{Novel} & \textbf{HM} \\
\midrule
CLIP        & 68.12 & 71.92 & 69.97 & CLIP        & 68.12 & 71.92 & 69.97 & CLIP        & 68.12 & 71.92 & 69.97 \\
PromptFL    & 69.55 & 69.01 & 69.28 & PromptFL    & 73.03 & 72.08 & 72.55 & PromptFL    & 68.07 & 64.24 & 66.10 \\
RPO & 65.96 & 70.55 & 68.17 & RPO & 66.80 & 71.50 & 69.07 & RPO & 68.75 & 71.13 & 69.92 \\
FedTPG      & 64.69 & 64.08 & 64.39 & FedTPG      & 68.86 & 67.47 & 68.16 & FedTPG      & 68.86 & 66.73 & 67.77 \\
TCP         & 73.82 & 72.47 & 73.13 & TCP         & 75.83 & 72.27 & 74.00 & TCP         & 76.09 & 73.00 & 74.51 \\
PromptFolio & 73.72 & 69.80 & 71.71 & PromptFolio & 76.56 & 70.12 & 73.20 & PromptFolio & 78.15 & 68.37 & 72.94 \\
DP-FPL & 72.13 & 70.65 & 71.38 & DP-FPL & 74.08 & 71.76 & 72.91 & DP-FPL & 73.24 & 68.16 & 70.61 \\
\hline
\methodname{}$_{P}$      & 76.88 & 68.53 & 72.47 & \methodname{}$_{P}$      & \textbf{79.10} & 68.75 & 73.56 & \methodname{}$_{P}$      & \textbf{78.89} & 69.38 & 73.83 \\
\methodname{}$_{T}$      & \textbf{77.78} & \textbf{74.62} & \textbf{76.17} & \methodname{}$_{T}$      & 78.15 & \textbf{75.79} & \textbf{76.95} & \methodname{}$_{T}$      & 78.04 & \textbf{76.58} & \textbf{77.31} \\
\bottomrule
\end{tabular}
}
\label{tab8}
\end{table*}
To evaluate the effectiveness of the proposed \methodname{} under different levels of local training effort, this section investigates its performance across varying local training epochs. Specifically, we consider three settings with local training epochs $\in \{1, 5, 20\}$, under a data heterogeneity level ($\beta = 0.5$) and with 10 clients participating in training. As shown in Table~\ref{tab7} and Table \ref{tab8}, both \methodname{}$_{P}$ and \methodname{}$_{T}$ consistently outperform the baselines across all datasets and training epochs. Notably, \methodname{}${T}$ delivers the best overall performance in terms of harmonic mean (HM) in most settings, which demonstrates superior generalization across both base and novel classes. On Flowers102 and UCF101, \methodname{}${T}$ shows strong robustness even when the number of local epochs increases. This highlights its ability to prevent overfitting to local distributions. Across datasets, both variants of \methodname{} remain stable and effective even as the local training dynamics change, outperforming methods that are more sensitive to overfitting or class imbalance under prolonged local updates.

\subsubsection{Sensitivity Analysis Regarding the Number of Prompt Tokens.}
\begin{table}[h]
\centering
\caption{Comparison of different methods under two settings (5×512 and 20×512) on Caltech101, DTD, and UCF101 datasets. Bold numbers indicate the best performance.}
\setlength{\tabcolsep}{4.5pt}
\renewcommand{\arraystretch}{1.15}
\begin{tabular}{lcccccccccccc}
\toprule
& \multicolumn{3}{c}{Caltech101} & \multicolumn{3}{c}{DTD} & \multicolumn{3}{c}{UCF101} \\
\cmidrule(lr){2-4} \cmidrule(lr){5-7} \cmidrule(lr){8-10}
& Base & Novel & HM & Base & Novel & HM & Base & Novel & HM \\
\midrule
\multicolumn{10}{l}{\textbf{5×512}} \\
PromptFL     & 88.41 & \textbf{93.97} & 91.11 & 59.25 & 52.77 & 55.83 & 76.78 & 69.91 & 73.18 \\
FedTPG       & 86.65 & 93.37 & 89.88 & 60.18 & 56.15 & 58.10 & 74.30 & 70.63 & 72.41 \\
PromptFolio  & 89.47 & 93.46 & 91.42 & \textbf{61.02} & \textbf{56.28} & 57.62 & 74.40 & \textbf{70.65} & 72.48 \\
GRP-NIAM\_P  & \textbf{89.90} & 93.37 & \textbf{91.60} & 60.96 & 55.95 & \textbf{59.12} & \textbf{77.98} & 70.39 & \textbf{73.99} \\
\midrule
\multicolumn{10}{l}{\textbf{20×512}} \\
PromptFL     & 88.91 & \textbf{94.31} & 91.53 & \textbf{62.50} & 52.53 & 57.08 & 76.83 & 69.12 & 72.77 \\
FedTPG       & 88.84 & 92.86 & 90.81 & 58.10 & 54.83 & 56.41 & 76.30 & 73.41 & 74.41 \\
PromptFolio  & 89.97 & 94.22 & 92.05 & 56.13 & 53.98 & 55.03 & 75.72 & 74.41 & 75.06 \\
GRP-NIAM\_P  & \textbf{90.02} & 94.28 & \textbf{92.10} & 61.43 & \textbf{56.23} & \textbf{58.71} & \textbf{77.23} & \textbf{74.58} & \textbf{75.88} \\
\bottomrule
\end{tabular}
\end{table}
 We have conducted additional sensitivity experiments regarding the number of prompt tokens. Specifically, we adjusted the prompt length to 5×512 and 20×512. The results show that GPR-NIAM remains robust across different prompt lengths, while consistently maintaining an advantage over all baselines in all case.

\subsubsection{Evaluating Method Scalability with Varying Client Numbers}
\begin{table*}[h]
\small
\centering
\caption{Performance comparison between \methodname{} and baselines on DTD and Flowers102 datasets under varying number of clients $K \in \{20, 30\}$ in the base-to-base/novel generalization.}
\label{tab9}
\renewcommand{\arraystretch}{1.0}
\resizebox{0.78\textwidth}{!}{
\begin{tabular}{lccc|lccc}
\toprule
\multicolumn{4}{c|}{\textbf{(a) DTD (K=20)}} & 
\multicolumn{4}{c}{\textbf{(b) DTD (K=30)}} \\
\hline
\textbf{Method} & \textbf{Base} & \textbf{Novel} & \textbf{HM} & 
\textbf{Method} & \textbf{Base} & \textbf{Novel} & \textbf{HM} \\
\midrule
CLIP        & 53.35 & 59.05 & 56.06 & CLIP        & 53.35 & 59.05 & 56.06 \\
PromptFL    & 54.16 & 54.22 & 54.19 & PromptFL    & 52.89 & 56.52 & 54.64 \\
RPO         & 53.24 & \textbf{60.02} & 56.42 & RPO         & 52.77 & \textbf{59.66} & 56.00 \\
FedTPG      & 57.83 & 54.93 & 56.34 & FedTPG      & 52.19 & 58.09 & 54.98 \\
TCP         & 58.21 & 55.93 & 57.04 & TCP         & 56.82 & 58.42 & 57.60 \\
PromptFolio & 58.18 & 52.91 & 55.41 & PromptFolio & 55.43 & 52.41 & 53.88 \\
DP-FPL      & 56.13 & 50.72 & 53.29 & DP-FPL      & 57.06 & 51.20 & 53.97 \\\hline
\methodname{}$_{P}$ & \textbf{63.31} & 52.29 & 57.27 & \methodname{}$_{P}$ & \textbf{60.64} & 51.93 & 55.95 \\
\methodname{}$_{T}$ & 62.38 & 53.86 & \textbf{57.81} & \methodname{}$_{T}$ & 60.41 & 56.52 & \textbf{58.40} \\
\midrule
\multicolumn{4}{c|}{\textbf{(d) Flowers102 (K=20)}} & 
\multicolumn{4}{c}{\textbf{(e) Flowers102 (K=30)}} \\
\hline
\textbf{Method} & \textbf{Base} & \textbf{Novel} & \textbf{HM} & 
\textbf{Method} & \textbf{Base} & \textbf{Novel} & \textbf{HM} \\
\midrule
CLIP        & 62.39 & 78.10 & 69.36 & CLIP        & 62.39 & 78.10 & 69.36 \\
PromptFL    & 74.34 & 74.10 & 74.22 & PromptFL    & 72.01 & 74.73 & 73.34 \\
RPO         & 71.72 & 76.84 & 74.19 & RPO         & 71.13 & 76.42 & 73.68 \\
FedTPG      & 74.63 & \textbf{78.31} & 76.43 & FedTPG      & 74.92 & 78.31 & 76.58 \\
TCP         & 75.21 & 77.68 & 76.43 & TCP         & 73.46 & 78.73 & 76.01 \\
PromptFolio & 75.51 & \textbf{78.31} & 76.88 & PromptFolio & 73.76 & 78.31 & 75.97 \\
DP-FPL      & 74.34 & 74.10 & 74.22 & DP-FPL      & 69.09 & 75.57 & 72.19 \\\hline
\methodname{}$_{P}$ & 82.79 & 77.05 & 79.82 & \methodname{}$_{P}$ & 81.63 & 78.31 & 79.93 \\
\methodname{}$_{T}$ & \textbf{83.96} & 77.05 & \textbf{80.36} & \methodname{}$_{T}$ & \textbf{82.50} & \textbf{78.94} & \textbf{80.68} \\
\bottomrule
\end{tabular}
}
\end{table*}

\begin{table*}[t]
\small
\centering
\caption{Performance comparison between \methodname{} and baselines on StanfordCars and UCF101 datasets under varying number of clients $K \in \{20, 30\}$ in the base-to-base/novel generalization.}
\label{tab10}
\renewcommand{\arraystretch}{1.0}
\resizebox{0.78\textwidth}{!}{
\begin{tabular}{lccc|lccc}
\toprule
\multicolumn{4}{c|}{\textbf{(g) StanfordCars (K=20)}} & 
\multicolumn{4}{c}{\textbf{(h) StanfordCars (K=30)}} \\
\hline
\textbf{Method} & \textbf{Base} & \textbf{Novel} & \textbf{HM} & 
\textbf{Method} & \textbf{Base} & \textbf{Novel} & \textbf{HM} \\
\midrule
CLIP        & 56.87 & 69.40 & 62.51 & CLIP        & 56.87 & 69.40 & 62.51 \\
PromptFL    & 59.82 & 67.03 & 63.22 & PromptFL    & 54.42 & 66.33 & 59.79 \\
RPO         & 57.22 & \textbf{69.65} & 62.83 & 
RPO         & 57.24 & \textbf{69.60} & 62.82 \\
FedTPG      & 59.24 & 66.73 & 62.76 & FedTPG      & 59.42 & 69.08 & 63.88 \\
TCP         & 58.77 & 67.78 & 62.95 & 
TCP         & 59.32 & 68.21 & 63.45 \\
PromptFolio & 60.91 & 65.78 & 63.25 & PromptFolio & 59.59 & 65.25 & 62.29 \\
DP-FPL      & 57.74 & 65.65 & 61.44 & 
DP-FPL      & 58.37 & 67.75 & 62.71 \\\hline
\methodname{}$_{P}$ & 60.04 & 68.23 & 63.87 & \methodname{}$_{P}$ & 60.09 & 67.40 & 63.54 \\
\methodname{}$_{T}$ & \textbf{61.02} & 69.40 & \textbf{64.94} & \methodname{}$_{T}$ & \textbf{60.14} & 69.38 & \textbf{64.43} \\
\midrule
\multicolumn{4}{c|}{\textbf{(j) UCF101 (K=20)}} & 
\multicolumn{4}{c}{\textbf{(k) UCF101 (K=30)}} \\
\hline
\textbf{Method} & \textbf{Base} & \textbf{Novel} & \textbf{HM} & 
\textbf{Method} & \textbf{Base} & \textbf{Novel} & \textbf{HM} \\
\midrule
CLIP        & 68.12 & 71.92 & 69.97 & CLIP        & 68.12 & 71.92 & 69.97 \\
PromptFL    & 75.67 & 68.75 & 72.04 & PromptFL    & 71.34 & 66.04 & 68.59 \\
RPO         & 66.80 & 71.34 & 69.00 &
RPO         & 66.75 & 71.39 & 68.99 \\
FedTPG      & 74.88 & 74.78 & 74.83 & FedTPG      & 76.09 & 74.62 & 75.35 \\
TCP         & 74.08 & 73.83 & 73.96 & 
TCP         & 73.50 & 75.58 & 74.53 \\
PromptFolio & 74.45 & 71.76 & 73.08 & PromptFolio & 77.09 & 74.04 & 75.54 \\
DP-FPL      & 75.51 & 69.75 & 72.52 &
DP-FPL      & 72.71 & 68.96 & 70.79 \\\hline
\methodname{}$_{P}$ & 76.41 & 69.01 & 72.52 & \methodname{}$_{P}$ & 76.62 & 72.50 & 74.50 \\
\methodname{}$_{T}$ & \textbf{77.04} & \textbf{76.21} & \textbf{76.62} & \methodname{}$_{T}$ & \textbf{77.62} & \textbf{76.05} & \textbf{76.83} \\
\bottomrule
\end{tabular}
}
\end{table*}
This section aims to evaluate the scalability of the proposed \methodname{} with respect to the number of participating clients. We consider two settings with $K \in \{20, 30\}$ under a heterogeneity level $\beta = 0.5$ and fixed local training epochs of 5. As shown in Table~\ref{tab9} and Table \ref{tab10}, both \methodname{}$_{P}$ and \methodname{}$_{T}$ consistently outperform all baselines across datasets and client configurations. \methodname{}$_{T}$ achieves the best overall performance on most datasets and demonstrates strong generalization to novel classes. For example, on Flowers102 and UCF101, it achieves the highest harmonic mean (HM) under both client settings. Meanwhile, \methodname{}$_{P}$ maintains strong performance on base classes. These results show that \methodname{} maintains high performance and stability even as the number of clients increases, demonstrating its scalability and effectiveness in more challenging cases.

\subsubsection{Evaluation of Module Contributions}
\begin{table*}[h]
\centering
\caption{Ablation study on the effectiveness of NIAM and CSCR modules under the base-to-base/novel generalization. Results are reported on four datasets with PromptFL as the baseline.}
\label{tab11}
\renewcommand{\arraystretch}{1.0}
\resizebox{0.78\textwidth}{!}{
\begin{tabular}{cccc|cccc}
\toprule
\multicolumn{4}{c|}{\textbf{(a) DTD}} & 
\multicolumn{4}{c}{\textbf{(b) Flowers102}} \\
\hline
\textbf{Method} & \textbf{Base} & \textbf{Novel} & \textbf{HM} & 
\textbf{Method} & \textbf{Base} & \textbf{Novel} & \textbf{HM} \\
\midrule
PromptFL    & 55.55 & 56.76 & 56.15 & PromptFL    & 78.42 & 76.00 & 77.19 \\
+NIAM & 55.78 & 54.73 & 55.25 & 
+NIAM & 76.13 & 77.78 & 76.94 \\
+NIAM+CSCR    & 68.51 & 52.65 & 59.54 & 
+NIAM+CSCR    & 84.83 & 78.31 & 81.44 \\
\midrule
\multicolumn{4}{c|}{\textbf{(c) StanfordCars}} & 
\multicolumn{4}{c}{\textbf{(d) UCF101}} \\
\hline
\textbf{Method} & \textbf{Base} & \textbf{Novel} & \textbf{HM} & 
\textbf{Method} & \textbf{Base} & \textbf{Novel} & \textbf{HM} \\
\midrule
PromptFL    & 60.39 & 67.35 & 63.68 & PromptFL    & 73.03 & 72.08 & 72.55 \\
+NIAM & 55.17 & 68.13 & 60.96 & 
+NIAM & 67.96 & 73.02 & 70.39 \\
+NIAM+CSCR    & 62.09 & 67.05 & 64.48 & 
+NIAM+CSCR    & 79.10 & 68.75 & 73.56 \\
\bottomrule
\end{tabular}
}
\end{table*}
We evaluate the contributions of the NIAM and CSCR modules by gradually adding them to the PromptFL baseline. Experiments are conducted on four datasets using 10 clients, with heterogeneity level $\beta=0.5$ and 5 local training epochs. As shown in Table \ref{tab11}, NIAM improves performance on novel classes by preserving generalized knowledge. CSCR enhances base class performance through visual supervision. Together, they deliver the best overall performance in terms of harmonic mean (HM), highlighting their complementary effects.

\subsubsection{Sensitive Analysis of Textual Prompt Template}
\begin{table}[h]
\small
\centering
\caption{Prompt template comparison on Caltech101, DTD, and UCF101. Best numbers in \textbf{bold}.}
\setlength{\tabcolsep}{5pt}
\renewcommand{\arraystretch}{1.15}
\begin{tabular}{lccccccccc}
\toprule
\multirow{2}{*}{Template} &
\multicolumn{3}{c}{Caltech101} &
\multicolumn{3}{c}{DTD} &
\multicolumn{3}{c}{UCF101} \\
\cmidrule(lr){2-4}\cmidrule(lr){5-7}\cmidrule(lr){8-10}
& Base & Novel & HM & Base & Novel & HM & Base & Novel & HM \\
\midrule
A photo of a [CLASS]      & 91.59 & 94.22 & \textbf{92.89} & 68.40 & 53.50 & 60.04 & 78.57 & \textbf{70.28} & 74.19 \\
A picture of a [CLASS]    & 91.43 & \textbf{94.34} & 92.86 & 67.37 & 52.83 & 59.22 & 78.61 & 69.54 & 73.79 \\
An image showing [CLASS]  & \textbf{91.52} & 93.88 & 92.68 & \textbf{68.57} & \textbf{54.12} & \textbf{60.49} & \textbf{78.77} & 70.14 & \textbf{74.20} \\
\bottomrule
\end{tabular}
\end{table}
The model's performance is not highly sensitive to the choice of textual prompt templates. 
Following prior works such as CLIP and PromptFL, we use ``\texttt{a photo of a [CLASS]}'' as the default template. 
Since our method mainly optimizes the learnable prompt tokens ($T_P$) while keeping the text tokens ($T_T$) fixed, 
the initial template primarily serves as a starting point for semantic alignment and has limited impact on the final performance. 
We also conducted a small-scale experiment comparing different templates (e.g., ``\texttt{a picture of a [CLASS]}'', 
``\texttt{an image showing [CLASS]}''), and the variation in Top-1 accuracy was less than 1\%, 
indicating that the \textbf{GPR-NIAM} framework is robust to the choice of templates.

\newpage

\section*{NeurIPS Paper Checklist}





\begin{enumerate}

\item {\bf Claims}
    \item[] Question: Do the main claims made in the abstract and introduction accurately reflect the paper's contributions and scope?
    \item[] Answer: \answerYes{} 
    \item[] Justification: This study outlines the research problem and its contributions in both the abstract and introduction.

\item {\bf Limitations}
    \item[] Question: Does the paper discuss the limitations of the work performed by the authors?
    \item[] Answer:  \answerYes{} 
    \item[] Justification: The limitations are discussed in the conclusion and future work sections.
    \item[] Guidelines:
    \begin{itemize}
        \item The answer NA means that the paper has no limitation while the answer No means that the paper has limitations, but those are not discussed in the paper. 
        \item The authors are encouraged to create a separate ''Limitations'' section in their paper.
        \item The paper should point out any strong assumptions and how robust the results are to violations of these assumptions (e.g., independence assumptions, noiseless settings, model well-specification, asymptotic approximations only holding locally). The authors should reflect on how these assumptions might be violated in practice and what the implications would be.
        \item The authors should reflect on the scope of the claims made, e.g., if the approach was only tested on a few datasets or with a few runs. In general, empirical results often depend on implicit assumptions, which should be articulated.
        \item The authors should reflect on the factors that influence the performance of the approach. For example, a facial recognition algorithm may perform poorly when image resolution is low or images are taken in low lighting. Or a speech-to-text system might not be used reliably to provide closed captions for online lectures because it fails to handle technical jargon.
        \item The authors should discuss the computational efficiency of the proposed algorithms and how they scale with dataset size.
        \item If applicable, the authors should discuss possible limitations of their approach to address problems of privacy and fairness.
        \item While the authors might fear that complete honesty about limitations might be used by reviewers as grounds for rejection, a worse outcome might be that reviewers discover limitations that aren't acknowledged in the paper. The authors should use their best judgment and recognize that individual actions in favor of transparency play an important role in developing norms that preserve the integrity of the community. Reviewers will be specifically instructed to not penalize honesty concerning limitations.
    \end{itemize}

\item {\bf Theory assumptions and proofs}
    \item[] Question: For each theoretical result, does the paper provide the full set of assumptions and a complete (and correct) proof?
    \item[] Answer: \answerNA{} 
    \item[] Justification: This paper does not involve theoretical assumptions.

    \item {\bf Experimental result reproducibility}
    \item[] Question: Does the paper fully disclose all the information needed to reproduce the main experimental results of the paper to the extent that it affects the main claims and/or conclusions of the paper (regardless of whether the code and data are provided or not)?
    \item[] Answer: \answerYes{} 
    \item[] Justification: This paper provides a detailed description of the experimental setup, including the range of parameter tuning, and the code will be uploaded as supplementary material.

\item {\bf Open access to data and code}
    \item[] Question: Does the paper provide open access to the data and code, with sufficient instructions to faithfully reproduce the main experimental results, as described in supplemental material?
    \item[] Answer: \answerYes{} 
    \item[] Justification: The code will be uploaded as supplementary material.

\item {\bf Experimental setting/details}
    \item[] Question: Does the paper specify all the training and test details (e.g., data splits, hyperparameters, how they were chosen, type of optimizer, etc.) necessary to understand the results?
    \item[] Answer: \answerYes{} 
    \item[] Justification: The experimental setup is clearly described in both the main experiment section and the appendix.

\item {\bf Experiment statistical significance}
    \item[] Question: Does the paper report error bars suitably and correctly defined or other appropriate information about the statistical significance of the experiments?
    \item[] Answer: \answerYes{} 
    \item[] Justification: This paper adopts evaluation metrics and assessment methods consistent with existing studies.
    \item[] Guidelines:
    \begin{itemize}
        \item The answer NA means that the paper does not include experiments.
        \item The authors should answer ''Yes'' if the results are accompanied by error bars, confidence intervals, or statistical significance tests, at least for the experiments that support the main claims of the paper.
        \item The factors of variability that the error bars are capturing should be clearly stated (for example, train/test split, initialization, random drawing of some parameter, or overall run with given experimental conditions).
        \item The method for calculating the error bars should be explained (closed form formula, call to a library function, bootstrap, etc.)
        \item The assumptions made should be given (e.g., Normally distributed errors).
        \item It should be clear whether the error bar is the standard deviation or the standard error of the mean.
        \item It is OK to report 1-sigma error bars, but one should state it. The authors should preferably report a 2-sigma error bar than state that they have a 96\% CI, if the hypothesis of Normality of errors is not verified.
        \item For asymmetric distributions, the authors should be careful not to show in tables or figures symmetric error bars that would yield results that are out of range (e.g. negative error rates).
        \item If error bars are reported in tables or plots, The authors should explain in the text how they were calculated and reference the corresponding figures or tables in the text.
    \end{itemize}

\item {\bf Experiments compute resources}
    \item[] Question: For each experiment, does the paper provide sufficient information on the computer resources (type of compute workers, memory, time of execution) needed to reproduce the experiments?
    \item[] Answer: \answerYes{} 
    \item[] Justification: The computational resources used are described in the experimental implementation details section.
    
\item {\bf Code of ethics}
    \item[] Question: Does the research conducted in the paper conform, in every respect, with the NeurIPS Code of Ethics \url{https://neurips.cc/public/EthicsGuidelines}?
    \item[] Answer: \answerYes{} 
    \item[] Justification: The research conducted in this paper fully conforms to the NeurIPS Code of Ethics in every respect.

\item {\bf Broader impacts}
    \item[] Question: Does the paper discuss both potential positive societal impacts and negative societal impacts of the work performed?
    \item[] Answer: \answerYes{} 
    \item[] Justification: As discussed in the introduction, the study substantially reduces communication costs by avoiding multiple rounds of exchange, which is advantageous for deployment under stringent latency or bandwidth limitations.
    
\item {\bf Safeguards}
    \item[] Question: Does the paper describe safeguards that have been put in place for responsible release of data or models that have a high risk for misuse (e.g., pretrained language models, image generators, or scraped datasets)?
    \item[] Answer: \answerNA{} 
    \item[] Justification: The paper utilizes the fully open-source CLIP model and publicly available benchmark datasets that are widely used in existing literature. Therefore, no additional safeguards are necessary for responsible release.

\item {\bf Licenses for existing assets}
    \item[] Question: Are the creators or original owners of assets (e.g., code, data, models), used in the paper, properly credited and are the license and terms of use explicitly mentioned and properly respected?
    \item[] Answer: \answerYes{} 
    \item[] Justification: the paper cites all relevant works, and all utilized assets are open-source and used in accordance with their licenses.

\item {\bf New assets}
    \item[] Question: Are new assets introduced in the paper well documented and is the documentation provided alongside the assets?
    \item[] Answer: \answerYes{} 
    \item[] Justification: The code developed for this work is included as anonymized supplementary material

\item {\bf Crowdsourcing and research with human subjects}
    \item[] Question: For crowdsourcing experiments and research with human subjects, does the paper include the full text of instructions given to participants and screenshots, if applicable, as well as details about compensation (if any)? 
    \item[] Answer: \answerNA{} 
    \item[] Justification: The paper does not involve crowdsourcing nor research with human subjects.

\item {\bf Institutional review board (IRB) approvals or equivalent for research with human subjects}
    \item[] Question: Does the paper describe potential risks incurred by study participants, whether such risks were disclosed to the subjects, and whether Institutional Review Board (IRB) approvals (or an equivalent approval/review based on the requirements of your country or institution) were obtained?
    \item[] Answer: \answerNA{} 
    \item[] Justification: The paper does not involve crowdsourcing nor research with human subjects.

\item {\bf Declaration of LLM usage}
    \item[] Question: Does the paper describe the usage of LLMs if it is an important, original, or non-standard component of the core methods in this research? Note that if the LLM is used only for writing, editing, or formatting purposes and does not impact the core methodology, scientific rigorousness, or originality of the research, declaration is not required.
    \item[] Answer: \answerNA{} 
    \item[] Justification: This paper only uses LLMs for grammar checking and editing.

\end{enumerate}

\end{document}